\documentclass[letterpaper, 10 pt, conference]{ieeeconf}  

\IEEEoverridecommandlockouts                              

\usepackage[table]{xcolor}
\usepackage{graphics} 
\usepackage{epsfig} 
\usepackage{graphicx}
\usepackage{amsmath}
\usepackage{booktabs}
\usepackage{algorithm}
\usepackage{algorithmic}
\usepackage{subfigure}
\usepackage{tikz}
\usepackage{adjustbox}
\usepackage{multirow}
\usepackage{bbold}
\usepackage{hyperref}

\makeatletter
\newcommand{\eqnum}{\refstepcounter{equation}\textup{\tagform@{\theequation}}}
\makeatother

\def\argmin{\mathop{\rm argmin}}

\title{\LARGE \bf
Unified Task and Motion Planning using Object-centric Abstractions of Motion Constraints
}

\author{Alejandro Agostini \and Justus Piater
\thanks{Alejandro Agostini and Justus Piater are with the Department of Computer Science, University of Innsbruck, 6020 Innsbruck,  Austria. {\tt\small \{alejandro.agostini, justus.piater\}@uibk.ac.at}. {\it Corresponding author: Alejandro Agostini}.
}
}

\begin{document}

\maketitle
\thispagestyle{empty}
\pagestyle{empty}

\begin{abstract}
In task and motion planning (TAMP), the ambiguity and underdetermination of abstract descriptions used by task planning methods make it difficult to characterize physical constraints needed to successfully execute a task. 
The usual approach is to overlook such constraints at task planning level and to implement expensive sub-symbolic geometric reasoning techniques that perform multiple calls on unfeasible actions, plan corrections, and re-planning until a feasible solution is found.
We propose an alternative TAMP approach that unifies task and motion planning into a single heuristic search. Our approach is based on an object-centric abstraction of motion constraints that permits leveraging the computational efficiency of off-the-shelf AI heuristic search to yield physically feasible plans. These plans can be directly transformed into object and motion parameters for task execution without the need of intensive sub-symbolic geometric reasoning. 
\end{abstract}

\section{INTRODUCTION}
Everyday human-like scenarios are highly unstructured and unpredictable. For a robot to operate autonomously in such scenarios, the traditional approach is to use artificial intelligence (AI) heuristic search approaches to automatically generate sequences of abstract instructions, called task plans, which define the steps to complete a task from a given initial situation. Robotic motion planning approaches, in turn, are used to transform each abstract instruction into robot motions for task execution. However, AI and robotic techniques are historically incompatible. They were conceived independently for different purposes, using different representations and search techniques. Combining them in a single robotic architecture is a great challenge.
The ambiguity and under-determination of symbolic representations make it difficult to characterize domain- and task-specific physical constraints relevant for generating realizable plans. The usual approach is to overlook such constraints at task planning and let sub-symbolic reasoning methods search for real-valued object and motion parameters that permit grounding symbolic actions considering the geometrical constraints in the scenario. 
Since searching strategies in real-valued spaces are computationally expensive, approximated solutions, such as sampling-based methods, are adopted \cite{dantam2018incremental, garrett2020pddlstream, garrett2021integrated}. 
Sampling-based approaches are based on trial-and-error strategies that require multiple calls on unfeasible actions in simulated scenarios until a solution, if any, is found. If no solution is found, task plan correction and re-planning mechanisms are triggered, starting another intensive trial-and-error process.

In this work we propose an alternative approach that blends task and motion planning into a single heuristic search using a common representation. We define an object-centric abstraction of motion constraints, such as grasping, placement, and kinematic constraints, compatible with both, task and motion planning. This representation allows to {\it move} motion planning into task planning, leveraging the computational efficiency of off-the-shelf AI heuristic search to yield feasible plans that can be directly translated into object and motion parameters for task execution, without the need of intensive sub-symbolic geometric reasoning. 

\subsection{Related Works}
Over the last year, several approaches combining task and motion planning for the robotic execution of tasks in unstructured scenarios have been proposed \cite{garrett2021integrated}.
One of the most extensively used approaches is to {\it hierarchically decompose} a task into several simpler sub-tasks that can be easily solved and executed \cite{kaelbling2011hierarchical,lallement2014hatp,colledanchise2017how}.
%
Kaelbling et al. \cite{kaelbling2011hierarchical} propose interleaving hierarchical task planning with plan execution mechanisms on relatively small sub-tasks that permits limiting the reasoning effort. The approach generates a global plan without checking in detail the forward progression of the effects of actions, 
focusing on the execution of the task at hand but at the expense of facing frequent planning impasses.

Some recent contributions propose {\it semantic representations of geometric constraints} to assess motion feasibility during task planning. 
Wells et al. \cite{wells2019learning} train a support vector machine to quickly classify motions as feasible or not feasible. This classification is associated to a generic proposition in the task planning domain that takes values true and false depending on the motion feasibility. The classifier has a relatively low accuracy due to the coarse granularity of semantic representations but reduces the effort of motion exploration. 
Bidot et al. \cite{bidot2017geometric} combine task and motion planning using a hybrid representation of symbolic and geometric states. Geometric states are used to check feasibility of symbolic actions in a plan. If an unfeasible action is found, a {\it geometric backtracking} is implemented to correct the plan.
This strategy requires intensive computations and several calls to motion planning on unfeasible actions to search for solutions in large object configuration spaces.
Dantam et al. \cite{dantam2018incremental} incorporate semantics descriptions of geometrical constraints to evaluate motion feasibility of single actions. The task planner adds and removes constraints incrementally while a {\it sampling-based motion planner} checks actions feasibility.  
In the same line, Garrett et al. \cite{garrett2020pddlstream} propose the PDDLstream approach, which includes functions called Streams that interface sampling-based procedures with task planning. 
All the previous approaches using symbolic representation of geometrical constraints requires expensive sub-symbolic mechanisms to assess motion feasibility outside task planning in a ``black-box" approach. Instead, our object-centric abstractions permit evaluating motion feasibility at task planning level, without the need of additional representations or sub-symbolic reasoning.

Another family of TAMP approaches {\it combine discrete search and optimal control} to search for solutions directly in the plan space, rather than in the state space, which permits better considering constraints across a plan. 
%
Lagriffoul et al. \cite{lagriffoul2016combining} use logic programming to find plans compatible with symbolic constraints as explanations of plan failures produced by collisions. 
Toussaint \cite{toussaint2015logic}, in turn, proposes an approach specially designed for creating pile of objects with stable configurations, where symbols are tailored to describe geometric and differential constrains for optimizing the entire plan execution. 
In the same vein, Fernandez et al. \cite{fernandez2017mixed} propose an optimization approach that scales better to longer horizon plans using heuristics that do not require time discretization.
These approaches require a model of the robot dynamics and intensive computations to find optimal solutions. 
Our approach, instead, does not require the robot dynamics and permits using the computational efficiency of off-the-shelf state-based planners to find solutions that jointly consider state and motion constraints.
Castaman et al. \cite{castaman2021receding} also borrows concepts from optimal control to propose a TAMP approach inspired by Model Predictive Control theory, where a task planner quickly generates constraint-agnostic sequence of actions and a sub-symbolic geometric reasoner searches for robot configurations in a finite horizon. The first action is executed and the entire process starts over again. The approach is computationally intensive, as it requires not only sub-symbolic geometric reasoning to find feasible motions, but also the computation of a TAMP solution at each time step. 

Object-centric representations are gaining attention to execute robotic actions thanks to their good generalization capabilities for manipulation planning
\cite{kroemer2019review,devin2018deep,wang2021dynamics,veerapaneni2019entity,garrett2021integrated,king2016rearrangement}, where object-centric descriptions are encoded as entities that generalize across objects presenting the same physical laws. 
King et al. \cite{king2016rearrangement} exploit the benefits of object-centric representations to define compatible robot-object configurations for the successful execution of high-level primitives (symbolic actions) in rearrangement tasks. However, they do not propose mechanisms to articulate object-centric motion planning with task planning for the generation of long-horizon manipulation plans. 
Exploiting object-centric representations for TAMP is the core idea of our previous approaches \cite{agostini2020perception, agostini2020manipulation}. 
In that contribution, object-centric predicates are used to assess geometrical plausibility of object configurations in states during the heuristic search of task planning.
Motion constraints, in turn, are learned from demonstration and encoded outside task planning in a hybrid symbol-signal representation call Action Context.
We propose a new TAMP approach built upon the concepts in \cite{agostini2020manipulation}.
Our approach {\it unifies task and motion planning} into a single heuristic search by defining object-centric abstractions of motion constraints, avoiding the need of external structures or learning from demonstration. 
In our previous contributions, geometrical constraints were encoded only for single object interactions (e.g. hand-object, object-support, etc.). 
Our new TAMP is able to encode {\it geometrical and motion constraints for multiple object interactions}, significantly scaling the applicability of our previous works.
The new TAMP approach also incorporates mechanisms to {\it dynamically generate tabletop parts} depending on object poses and available spaces. This permits randomly placing objects on the table as well as robustly reacting to disturbances, contrary to our previous approaches that restricted object positions to handcrafted tabletop parts in semi-structured scenarios.

\section{BASIC ELEMENTS}
\label{sec:basicnotation}
\subsection{Object-centric Abstractions}
\label{sec:object-centric abstractions}
We define a set of symbolic variables that will characterize constraints for TAMP in terms of functional parts of objects. In this work, we consider tasks where object functionality can be characterized using parts obtained from objects' bounding boxes and orientation in an object-centric approach: {\tt on} for the top of the object, {\tt under} for the bottom, and {\tt right}, {\tt left}, {\tt front}, and {\tt back} for the rest of the sides of the bounding box. We also consider the inside of an object as a part denoted by {\tt in}. 
Using these parts, we define constraint-related symbolic variables such as {\tt ?o-h-f1} (short for relations between a grasped object {\tt o}, and the Finger 1 of the robot hand, {\tt h-f1}) to characterize grasping constraints, {\tt ?o-loc} (short for object-support relations) to characterize placement constraints and so on, where the question mark denotes it is a variable. 
\subsection{Task Planning}
\label{sec:task_planning}
For task planning, we will use the Planning Domain Definition Language (PDDL) notation \cite{mcdermott1998pddl} and define a set of objects (e.g. {\small \tt cup}, {\small \tt table}) and a set of predicates coding object relations and properties (e.g. {\small \tt on cup table}). Predicates are logical functions that takes value {\small \tt true} or {\small \tt false}. The set of predicates describing a particular scenario defines a \textit{symbolic state} $s$.
We define a set of planning operators (POs) represented in the traditional precondition-effect manner. The precondition part comprises the predicates that change by the execution of the PO, as well as those predicates that are necessary for these changes to occur. The effect part describes the changes in the symbolic state after the PO execution.
The action is the name of the PO and consists of a declarative description of an action and may contain parameters to ground the predicates in the precondition and effect parts.
In task planning \cite{ghallab2004automated}, the planner receives the description of the \textit{initial state}, ${\tt s_{ini}}$, and a \textit{goal} description, {\tt g}, consisting of a set of grounded predicates that should be observed after task execution. With these elements, the planner searches for a sequence of actions called {\it plan} that would permit producing changes in ${\tt s_{ini}}$ necessary to obtain the goal {\tt g}.
\subsection{Physical State}
\label{sec:physical_state}
\begin{figure}
	\begin{center}
        \includegraphics[width=0.35\columnwidth]{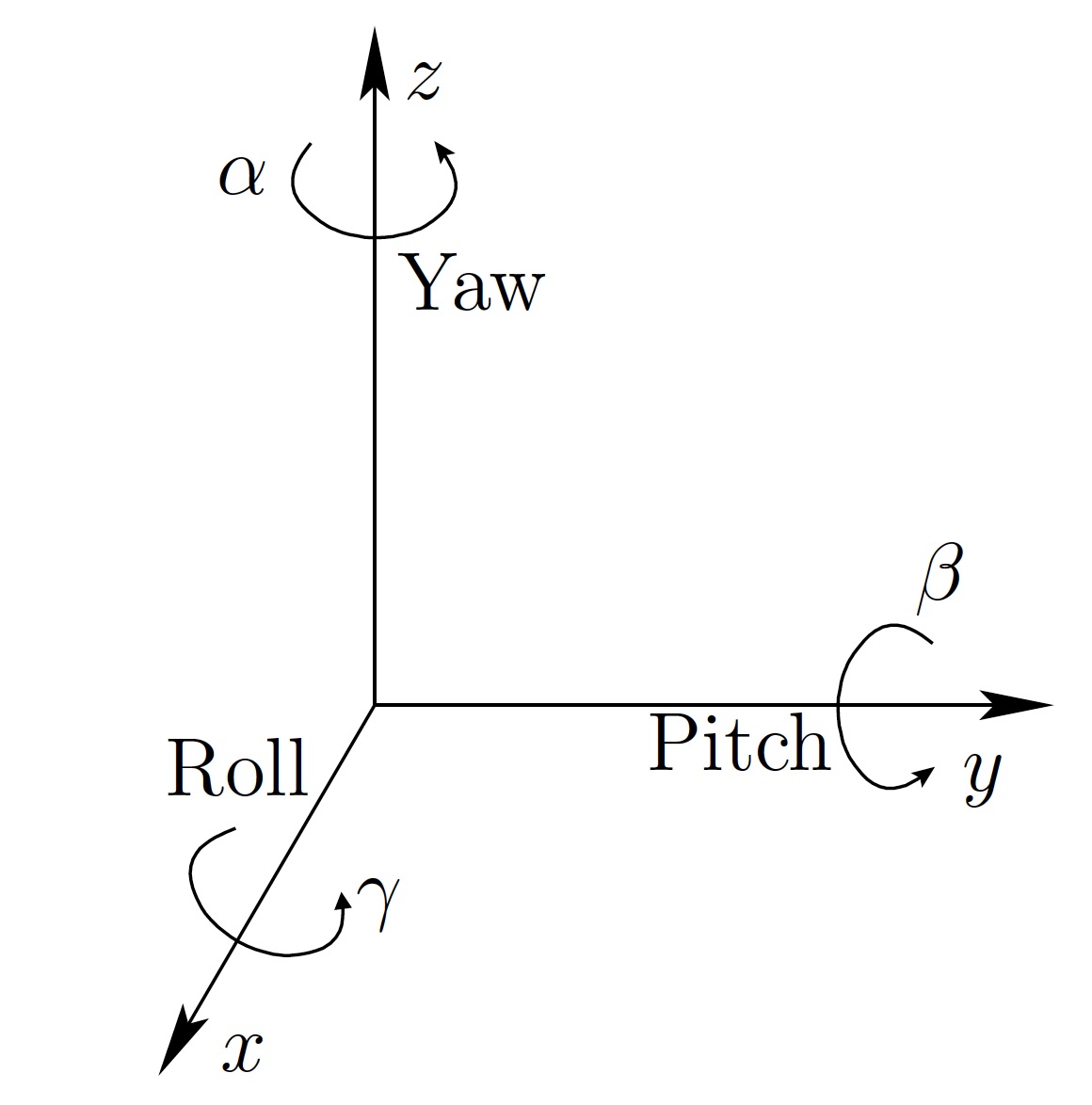}
		\caption{3D rotation in terms of roll, pitch, and yaw. Source \cite{lavalle2006planning}. 
		}
		\label{fig:roll_pitch_yaw}
	\end{center}
\end{figure}

We define a {\it physical state} of object $i$ as $z_i = \{ \:_r\xi_i, \: \Delta_i \}$, where $_r\xi_i = \{\:_rp_i, \: _rw_i\}$ is the pose of the object with respect to reference frame $\{r\}$, $_rp_i$ and $_rw_i$ are the position and orientation, respectively, and $\Delta_i = \{ \Delta^x_i, \Delta^y_i, \Delta^z_i \}$ is the size of the bounding box of the object in the Cartesian space. 
The orientation $_rw_i$ is represented using the roll ($\gamma$), pitch ($\beta$), and yaw ($\alpha$) terminology \cite{lavalle2006planning}, which represents rotations around the $x$, $y$, and $z$ axes, respectively (Fig. \ref{fig:roll_pitch_yaw}). 
There are different ways of obtaining the orientation of an object depending on the sequence of rotation around each axis. In this work we consider the "XYZ" sequence \cite{lavalle2006planning}.
The roll, pitch, and yaw can be used to represent the orientation of an object in matrix notation, $R(\alpha, \beta, \gamma) = R^z(\alpha) \: R^y(\beta) \: R^x(\gamma)$, where
\begin{equation}
\nonumber
R^z(\alpha) = 
\begin{bmatrix}
cos(\alpha) & -sin(\alpha) & 0\\
sin(\alpha) & cos(\alpha) & 0\\
0 & 0 & 1
\end{bmatrix},
\end{equation}
\begin{equation}
\nonumber
R^y(\beta) = 
\begin{bmatrix}
cos(\beta) & 0 & sin(\beta)\\
0 & 1 & 0\\
-sin(\beta) & 0 & cos(\beta)
\end{bmatrix},
\end{equation}
\begin{equation}
\nonumber
R^x(\gamma) = 
\begin{bmatrix}
1 & 0 & 0\\
0 & cos(\gamma) & -sin(\gamma)\\
0 & sin(\gamma) & cos(\gamma)
\end{bmatrix}.
\end{equation}
\noindent In turn, given a rotation matrix $R(\alpha, \beta, \gamma)$, we can obtain the roll, pitch and roll as \cite{lavalle2006planning}:
\begin{align}
\nonumber
\alpha &= \tan^{-1} (r_{21}/r_{11}), \\
\nonumber
\beta &= \tan^{-1} (-r_{31} / \sqrt{r_{32}^2 + r_{33}^2}), \\
\gamma &= \tan^{-1} (r_{32}/r_{33})
\label{eq:r2w}
\end{align}
\noindent where $r_{ij}$ are elements of the matrix. 

\subsection{Motion Planning}
The motion planning problem consists in finding trajectories $\tau:[0,1] \rightarrow Q$, where $Q$ is the robot configuration space, such that $\forall \lambda \in [0,1], \tau(\lambda) \in Z^F$, where $Z^F$ are valid configurations along the trajectory that satisfy the constraints $F: Z \rightarrow \{0,1\}$ \cite{garrett2021integrated}.
Similarly to task planning, motion planning defines an initial state $_r\xi_h^{ini}=\tau(0)$ representing the initial robot configuration, and a goal state $_r\xi_h^{goal}=\tau(1)$ that belongs to the set of possible goal configurations of the robot.
{\it Multi-modal} motion planning extends the motion planning problem to sequences of actions that require different motion {\it modes}, where each mode $\sigma$ is characterized by its own set of constraints $F_{\sigma}$ \cite{garrett2021integrated}.
\subsection{TAMP Action Template}
\label{sec:tamp_action_template}
\begin{figure}
	\begin{center}
        \includegraphics[width=1\columnwidth]{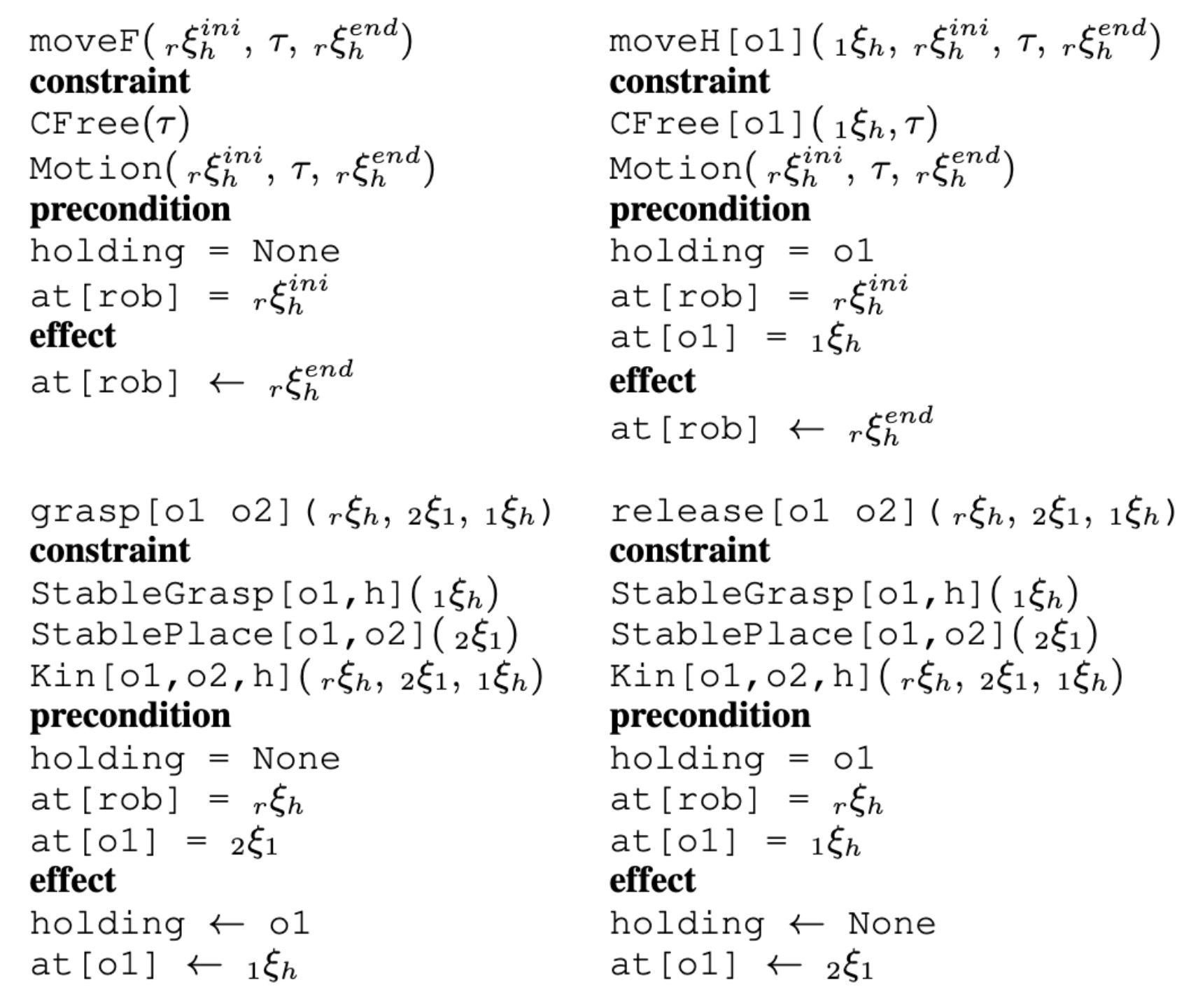}
		\caption{TAMP templates for pick-and-place actions adapted from \cite{garrett2021integrated}.
		}
		\label{fig:tamp_templates}
	\end{center}
\end{figure}

In this section, we briefly introduce generic TAMP action templates for pick-and-place following the terminology proposed by Garrett et al. \cite{garrett2021integrated}.
These action templates combine symbolic and real-valued parameters as required to solve hybrid constraint satisfaction problem (H-CSP) for TAMP \cite{garrett2021integrated}.
We define four TAMP action templates (Fig. \ref{fig:tamp_templates}). Two templates for moving the robot hand with and without holding an object and two templates for grasping and releasing objects.

The moving action {\tt moveF} ("move Free") corresponds to a motion in a {\it transit} mode, i.e. the robot moves the hand without holding an object along a collision-free trajectory $\tau$, where collisions are assessed through the constraint {\tt CFree}$(\tau)$. In this action, we include a motion constraint {\tt Motion}$(\: _r\xi_h^{ini},\: \tau, \:_r\xi_h^{end})$ that defines the relation between the initial $_r\xi_h^{ini}$ and final $_r\xi_h^{end}$ poses of the hand with the trajectory $\tau$. 
The other moving action corresponds to a {\it transfer} mode, ({\tt moveH}, "move Holding"), i.e. the robot moves the hand while holding an object, say {\tt o1}, along a collision-free trajectory $\tau$, with collision constraint {\tt CFree[o1]}$(\:_1\xi_h,\tau)$ and motion constraint {\tt Motion}$(\: _r\xi_h^{ini},\: \tau, \:_r\xi_h^{end})$.

In addition to the moving actions, we define action templates {\tt grasp} and {\tt release} for grasping and releasing objects, respectively.
The action {\tt grasp[o1 o2]}($\:_r\xi_h, \:_2\xi_1, \:_1\xi_h$) defines a grasping of an object {\tt o1} that is placed on a support {\tt o2}. A successful grasping takes place if the object is stably grasped and if the grasped object is stably placed on a support. These constraints are assessed as {\tt StableGrasp[o1,h]}$(\:_{1}\xi_{h})$ and {\tt StablePlace[o1,o2]}$(\:_2\xi_1)$, respectively, where $_{1}\xi_{h}$ are legal grasping poses of the hand {\tt h} in the reference frame of object {\tt o1}, and $_{2}\xi_{1}$ are legal placement poses of {\tt o1} with respect to a support object {\tt o2}.
The {\tt release[o1 o2]}($\:_r\xi_h, \:_2\xi_1, \:_1\xi_h$) action, in turn, defines stable releasing of object {\tt o1} on a support object {\tt o2}, assuming that the object is stably grasped and that the placement pose is stable, represented as before with the constraints {\tt StableGrasp[o1,h]}$(\:_{1}\xi_{h})$ and {\tt StablePlace[o1,o2]}$(\:_2\xi_1)$, respectively.
Finally, a kinematic constraint {\tt Kin[o1,o2,h]}$(\:_r\xi_h,\:_2\xi_1,\:_1\xi_h)$ defines the relation between the hand position in the global reference frame $\:_r\xi_h$ for specific grasping and placement configurations.

\section{U-TAMP}
\label{sec:U-TAMP}
In this section, we describe our unified task and motion planning (U-TAMP) approach using object-centric abstractions of the motion constraints presented in Sec. \ref{sec:tamp_action_template}.
We first introduce, in Sec. \ref{sec:U-TAMP Variables}, the variables that will be used to characterize motion constraints using object-centric abstractions and how to map them to real-valued object and motion parameters. 
Then, in Sec. \ref{sec:U-TAMP Constraints}, we present how motion constraints are described using predicates in terms of object-centric abstractions. 
Sec. \ref{sec:U-TAMP Actions} introduces task planning action templates encoding the object-centric constraints of Sec. \ref{sec:U-TAMP Constraints} to consider them during the heuristic search of task planning so as to yield task plans that satisfy these constraints.
We conclude with a description of the perception and execution mechanisms for transforming sensing parameters   into object-centric abstractions for task planning (Sec. \ref{sec:U-TAMP Perception}) and how U-TAMP actions are transformed into motion parameters for task execution (Sec. \ref{sec:U-TAMP Execution}).

\subsection{U-TAMP Variables}
\label{sec:U-TAMP Variables}
In this section, we describe the basic mechanisms for mapping real-valued object and motion parameters into object-centric abstractions for the assessment of grasping, placement, and kinematic constraints for the actions {\tt grasp} and {\tt release} in Fig. \ref{fig:tamp_templates}. Constraints for the actions {\tt moveF} and {\tt moveH} are explained together with the mechanisms for task execution in Sec. \ref{sec:U-TAMP Execution}.

\subsubsection{Grasping}
\label{sec:legal_grasping}
\begin{figure}[!h]
 \begin{center}
    \includegraphics[width=0.8\columnwidth]{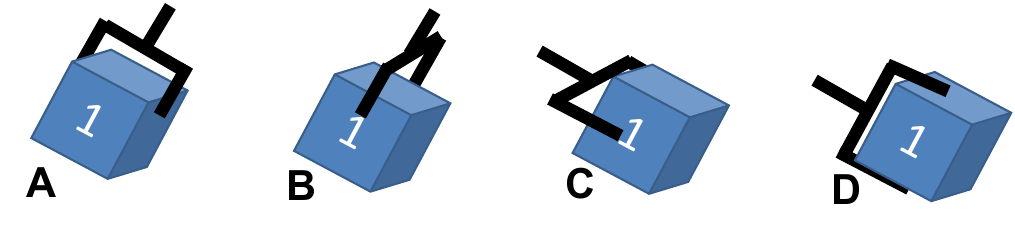}
    		\caption{Schema of hand-object configurations for four different grasp types. 
    		}
    \label{fig:example_grasp_types}
 \end{center}
\end{figure}

In order to define legal grasping configurations in terms of interactions between functional parts of the grasped object and of the robot hand, we use the symbolic variables {\tt ?o1-h-p}, {\tt ?o1-h-f1}, and {\tt ?o1-h-f2}, where {\tt ?o1-h-p} represents the part of object {\tt o1} interacting with the palm of the hand, {\tt ?o1-h-f1} represents the part of the object interacting with the Finger 1 of the hand, and {\tt ?o1-h-f2} is the part of the object interacting with the Finger 2 of the hand.
All the object parts are assumed to be graspable, i.e., to have a grasping affordance.
By assigning values to these variables, we can unequivocally define 24 different hand-object poses $_1\xi_h = \{\:_1p_h,\:_1w_h\}$ for the evaluation of the grasp constraint {\tt StableGrasp[o1]}$(\:_{1}\xi_{h})$, four poses for each functional part of {\tt o1}.
Fig. \ref{fig:example_grasp_types} presents four example hand-object configurations. The specific values for the arguments {\tt ?o1-h-p}, {\tt ?o1-h-f1}, {\tt ?o1-h-f2} are shown in Table \ref{table:oRh}.  

Relative hand-object positions, $_1p_h$, are obtained from the value of the argument {\tt ?o1-h-p}, which indicates the surface of the object's part (side of bounding box) in contact with the palm of the hand. The position of the hand is calculated as the centroid of such surface, obtained from the size of the bounding box $\Delta$, as shown in Table \ref{table:oph}. For example, the relative hand-object position when the object is grasped from its top ({\tt ?o1-h-p} = {\tt on}) is given by the centroid of the top surface of the bounding box in the reference frame of {\tt o1}, $(0,0,\Delta^z_1 / 2)$. 
\begin{table}[h!]
\caption{Centroid of parts corresponding to sides of the bounding box of an object {\tt o1} calculated using $\Delta_1$.}
{\small
\begin{center}
\resizebox{0.7\columnwidth}{!}{
\begin{tabular}{llll}
Part & $_1x_h$ & $_1y_h$ & $_1z_h$ \\
\hline
{\tt on} & 0 & 0 & $\Delta^z_1/2$ \\
{\tt under} & 0 & 0 & $-\Delta^z_1/2$ \\
{\tt left} & 0 & $\Delta^y_1/2$ & 0 \\
{\tt right} & 0 & $-\Delta^y_1/2$ & 0 \\
{\tt front} & $\Delta^x_1/2$ & 0 & 0 \\
{\tt back} & $-\Delta^x_1/2$ & 0 & 0 \\
\end{tabular}}
\end{center}}
\label{table:oph}
\end{table}

Relative hand-object orientations, $_1w_h$, in turn, are obtained from the surfaces of the parts of {\tt o1} in contact with the hand when object is grasped, represented by the variables {\tt ?o1-h-p}, {\tt ?o1-h-f1}, {\tt ?o1-h-f2}. There is a unique hand-object orientation for each of the 24 possible configurations defined by these variables.
For example, the orientation for the configuration {\tt ?o1-h-p} = {\tt on}, {\tt ?o1-h-f1} = {\tt back}, and {\tt ?o1-h-f2} = {\tt front} (Fig. \ref{fig:example_grasp_types}B) can be obtained by, first, rotating the hand by $\pi$ around the $x$ axis of the object's reference frame ($_1\gamma_h = \pi$), no rotation around the $y$ axis ($_1\beta_h = 0$), and a clockwise rotation by $\pi/2$ around the object's $z$ axis ($_1\alpha_h = -\pi/2$).
Table \ref{table:oRh} presents the roll, pitch, and yaw rotations of the hand with respect to the object's reference frame for the four types of grasping in Fig. \ref{fig:example_grasp_types}.
\begin{table}[h!]
\caption{Yaw, pitch, and roll rotations of the hand with respect to the object reference frame for the examples cases in Fig. \ref{fig:example_grasp_types}.}
{\small
\begin{center}
\resizebox{\columnwidth}{!}{
\begin{tabular}{lllllll}
&{\tt ?o1-h-p} & {\tt ?o1-h-f1} & {\tt ?o1-h-f2} & $_1\gamma_h$ & $_1\beta_h$ & $_1\alpha_h$ \\
\hline
A & {\tt on} & {\tt left} & {\tt right} & $\pi$ & $0$ & $0$ \\
B & {\tt on} & {\tt back} & {\tt front} & $\pi$ &$0$ & $-\pi/2$ \\
C & {\tt right} & {\tt back} & {\tt front} & $-\pi/2$ & $0$ & $0$ \\
B & {\tt right} & {\tt under} & {\tt on} & $-\pi$ & $0$ & $-\pi/2$ \\
\end{tabular}}
\end{center}}
\label{table:oRh}
\end{table}

\subsubsection{Placement}
\label{sec:legal_placement}
\begin{figure}
	\begin{center}
        \includegraphics[width=0.8\columnwidth]{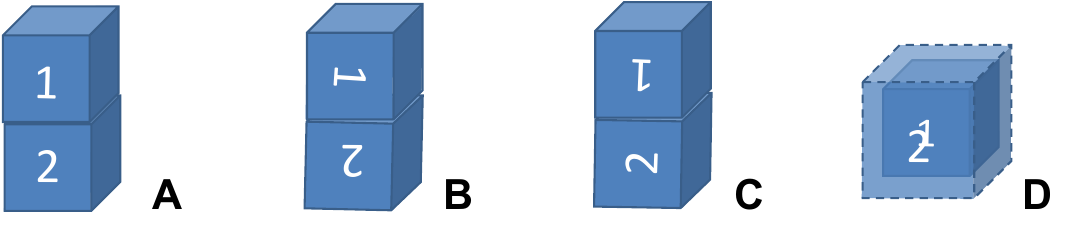}
		\caption{Example of four placement types, as described in Table \ref{table:example_place_types}. 
		}
		\label{fig:example_place_types}
	\end{center}
\end{figure}

Legal placement configurations $_2\xi_1$ are defined by interacting parts of an object and its support. We define the symbolic variables {\tt ?o1-o2} and {\tt ?o2-o1}, representing the part of the placed object {\tt o1} interacting with the support object {\tt o2}, and the part of support object {\tt o2} interacting with the placed object {\tt o1}, respectively, where part {\tt ?o2-o1} of object {\tt o2} is assumed to be able to support part {\tt ?o1-o2} of object {\tt o1}, i.e., to provide a support affordance for that part.
By assigning values to {\small \tt ?o2-o1} and {\small \tt ?o1-o2}, we can define 36 placement configurations, each of them having a unique object-support pose $_2\xi_1$.
Fig. \ref{fig:example_place_types} shows four examples placement configurations while Table \ref{table:example_place_types} depicts position and orientation for each of these examples. 
For instance, if the bottom surface of {\tt o1} is placed on the top surface of the support object {\tt o2} (Fig. \ref{fig:example_place_types}A), we have {\tt ?o1-o2} = {\tt under} and {\tt ?o2-o1} = {\tt on}. In this configuration, the relative position $_2p_1$ of object {\tt o1} in the reference frame of {\tt o2} is calculated as $_2p_1 = (0,0,\Delta^z_2 / 2 + \Delta^z_1 / 2)$, where $\Delta^z_2$ and $\Delta^z_1$ are the lengths of the bounding boxes of {\tt o1} and {\tt o2} along the $z$ axis, respectively. In this case, both objects are aligned, i.e. $_2w_1 = (0,0,0)$.
We also consider consider placement configurations defined by an object inside a (container) space (Fig. \ref{fig:example_place_types}D, see also Table \ref{table:example_place_types}), where the container is an empty space represented as a hollow bounding box. This configuration is represented as {\small \tt ?o2-o1 = in} and {\small \tt ?o1-o2 = in} and is used to characterize {\it placed-in} cases, as develop in Sec. \ref{sec:object-container tasks}, where both, the object and its container space, are assumed to be aligned.
\begin{table}[h!]
\caption{Relative poses for the example placement types in Fig. \ref{fig:example_place_types}.}
{\small
\begin{center}
\resizebox{\columnwidth}{!}{
\begin{tabular}{lllllllll}
&{\tt ?o1-o2} & {\tt ?o2-o1} & $_2x_1$ & $_2y_1$ & $_2z_1$ &$_2\gamma_1$ & $_2\beta_1$ & $_2\alpha_1$ \\
\hline
A & {\tt under} & {\tt on} & $0$ & $0$ & $(\Delta^{z}_2 + \Delta^{z}_1) / 2$ & $0$ & $0$ & $0$ \\
B & {\tt left} & {\tt under} & $0$ & $0$ & $(-\Delta^{z}_2 - \Delta^{y}_1) / 2$ & $\pi / 2$ & $0$ & $0$ \\
C & {\tt on} & {\tt left} & $0$ & $(\Delta^{y}_2 + \Delta^{z}_1) / 2$ & $0$ & $\pi / 2$ & $0$ & $0$ \\
D & {\tt in} & {\tt in} & $0$ & $0$ & $0$ & $0$ & $0$ & $0$ \\
\end{tabular}}
\end{center}}
\label{table:example_place_types}
\end{table}
\subsubsection{U-TAMP Kinematic}
\label{sec:legal_kinematics}
\begin{figure}
	\begin{center}
        \includegraphics[width=0.8\columnwidth]{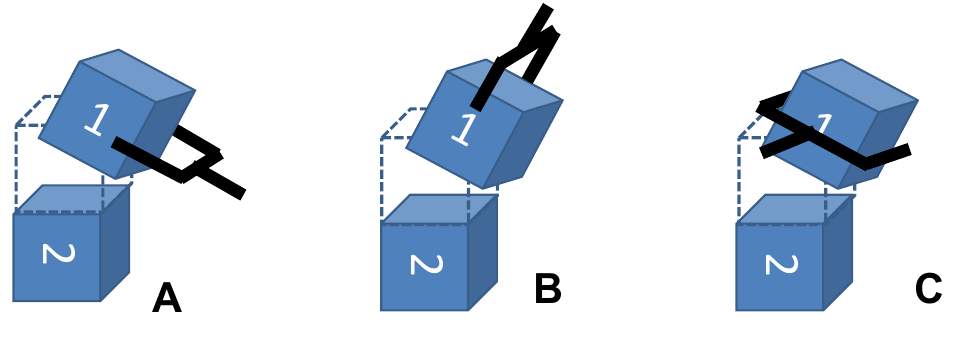}
		\caption{Example hand-object and object-support configurations in the kinematics constraint. 
		}
		\label{fig:example_kin_place}
	\end{center}
\end{figure}

Given the grasping $_1\xi_h$ (Sec. \ref{sec:legal_grasping}) and a placement $_2\xi_1$ (Sec. \ref{sec:legal_placement}) configurations, we can easily derive the 
hand pose $_r\xi_h = \{\:_rp_h,\:_rR_h\}$ \footnote{Here we use the matrix notation $_rR_h$ instead of the yaw-pitch-roll notation $_rw_h$ to represent the hand rotation (see Sec. \ref{sec:physical_state}).} with respect to the global reference frame, $\{r\}$, from the absolute pose of the grasped object {\tt o1} as
\begin{equation}
    _rp_h =\:_rp_1 + \:_rR_1 \:_1p_h,
    \label{eq:_rp_h}
\end{equation}
\noindent and 
\begin{equation}
    _rR_h = \:_rR_1 \:_1R_h,
    \label{eq:_rR_h}
\end{equation}
\noindent where $_1p_h$ is the position of the hand with respect to the reference frame of object {\tt o1}, and $_rp_1$ and $_rR_1$ are the absolute position and orientation (in matrix notation) of the grasped object {\tt o1}, respectively. 
In the case of picking actions, the pose of object {\tt o1} can be obtained directly from sensing mechanisms. In the case of placing actions, on the other hand, it can be derived from the absolute pose of the support object {\tt o2} as
\begin{equation}
    _rp_1 =\:_rp_2 + \:_rR_2 \:_2p_1,
    \label{eq:_rp_1}
\end{equation}
\noindent and
\begin{equation}
    \:_rR_1 = \:_rR_2 \:_2R_1,
    \label{eq:_rR_1}
\end{equation}
\noindent where $\:_rp_2$ and $\:_rR_2$ are the position and orientation of object {\tt o2} in the global reference frame and $_2p_1$ and $_2R_1$ are obtained as explained in Sec. \ref{sec:legal_placement}.
Fig. \ref{fig:example_kin_place} illustrates three example configurations of the hand, grasped object, and its support.
In this manner, if the grasp and placement constraints are fulfilled, i.e. {\tt StableGrasp[o1,h]}$(\:_1\xi_h) = 1$ and {\tt StablePlace[o1,o2]}$(\:_2\xi_1) = 1$, we can calculate the values of $_r\xi_h$ satisfying the constraint {\tt Kin[o1,o2,h]}$(\:_r\xi_h,\:_2\xi_1,\:_1\xi_h)$ through equations \eqref{eq:_rp_h} and \eqref{eq:_rR_h}. 
\subsection{U-TAMP Constraints}
\label{sec:U-TAMP Constraints}
In the previous sections, we have defined legal grasping and placement poses in terms of functional object parts. In this section we describe how to assess if these poses satisfy {\it task-dependent} TAMP constraints using predicates and object-centric abstractions (Sec. \ref{sec:object-centric abstractions}).
Our aim is to evaluate, on the one hand, which legal grasping configurations $_1\xi_h$, or equivalently, which values of {\tt ?o1-h-p}, {\tt ?o1-h-f1}, and {\tt ?o1-h-f2} (Sec. \ref{sec:legal_grasping}) satisfy the constraint {\tt StableGrasp[o1,h]}$(\:_{1}\xi_{h})$. On the other hand, which legal placement configurations $_2\xi_1$, or equivalently, which values of {\tt ?o1-o2} and {\tt ?o2-o1} (Sec. \ref{sec:legal_placement}) satisfy the constraint {\tt StablePlace[o1,o2]}$(\:_2\xi_1)$. Note that, once the values $_1\xi_h$ and $_2\xi_1$ satisfying constraints {\tt StableGrasp} and {\tt StablePlace} are found, it is possible to deduce the values $_r\xi_h$ so that {\tt Kin[o1,o2,h]}$(\:_r\xi_h,\:_2\xi_1,\:_1\xi_h) = 1$, as explained in Sec. \ref{sec:legal_kinematics}.

We define TAMP constraints for two types of pick-and-place tasks. 
The first type of tasks focused on interactions between a grasped object and its support, which we denote {\it object-support} tasks. Quite traditionally, these tasks requires reasoning about how to allocate objects on support surfaces to complete a goal.
The second type of tasks focuses on interactions between a manipulated object and its container. These tasks, denoted as {\it object-container} tasks, permit considering interactions between an object with many other objects adjacent to it.
\subsubsection{Object-support Tasks}
\label{sec:object-support tasks}
%
To characterize legal grasping configurations, we define the predicate {\tt isgrasp ?o1-h-p ?o1-h-f1 ?o1-h-f2}. By assigning values to the variables {\tt ?o1-h-p}, {\tt ?o1-h-f1}, and {\tt ?o1-h-f2}, the predicate takes value {\tt true} if the assigned values correspond to a legal grasping, e.g. {\tt isgrasp on left right} and value {\tt false} otherwise. 
Available grasping configurations in picking action are those where the sides of the bounding box of {\tt o1} corresponding to a legal grasping are clear. This is represented by the predicates predicates {\tt ?o1-h-p ?o1 air}, {\tt ?o1-h-f1 ?o1 air}, and {\tt ?o1-h-f2 ?o1 air} taking value {\tt true}, where {\tt air} is a virtual object denoting an empty space, in the same vein as in our previous work \cite{agostini2020manipulation}. For example, the grasping configuration {\tt isgrasp on left right} is available if {\tt on o1 air}, {\tt left o1 air}, and {\tt right o1 air} are {\tt true}.
In addition to constraints for evaluating legal grasping configurations, we define constraints that help selecting grasping configurations with plausible kinematic. To this end, we define the object-centric abstraction {\tt ?o1-base}, representing the side of the bounding box of object {\tt o1} that is the closest to the base of the robot. Using this variable, we define the predicate {\tt base ?o1 ?o1-base} that takes value {\tt true} if {\tt ?o1-base} is the surface closest to the robot's base. We restrict grasping configurations to those where the palm of the hand is not interacting with the part of object {\tt o1} which is in the opposite side of {\tt ?o1-base}. This is done by requiring the predicate {\tt not(isopposite ?o1-base ?o1-h-p)} to be {\tt true}.

\begin{figure}[h!]
	\begin{center}
        \includegraphics[width=1\columnwidth]{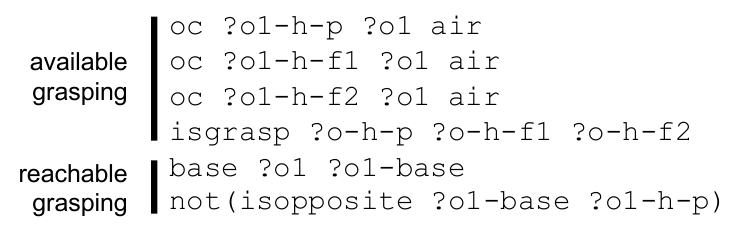}
		\caption{Predicates defining grasping constraints for object-support pick-and-place tasks.
		}
		\label{fig:grasp_constraint_pddl}
	\end{center}
\end{figure}

Fig. \ref{fig:grasp_constraint_pddl} compiles the predicates defining the grasping constraints for picking actions in object-support pick-and-place tasks. Grasping configurations {\tt ?o1-h-p}, {\tt ?o1-h-f1}, and {\tt ?o1-h-f2} satisfying these constraints define the set of hand-object poses $_1\xi_h$ so that {\tt StableGrasp[o1,h]}$(\:_1\xi_h) = 1$.

We define predicates to characterize placement constraints in placing actions in terms of object-support interactions {\small \tt ?o2-o1} and {\small \tt ?o1-o2}. 
Provided that any two parts of the placed object {\tt o1} and its support {\tt o2} define legal placement configurations, available placements are evaluated through the predicates {\tt ?o1-o2 ?o1 air} and {\tt ?o2-o1 ?o1 air}, which check if the parts of {\tt o1} and {\tt o2} that will be interact after a placing action are clear. 
However, stable placements are only possible if part {\tt ?o2-o1} of the support object {\tt o2} is able to steadily support {\tt o1}. To this end, we use the object-centric abstraction {\tt ?o2-force} that represents the part of the support object {\tt o2} that is able to support other objects given the environment balance of forces. For example, in standard settings where the gravity force prevails, this surface will be the upmost surface of {\tt o2}. Using this variable, we define the  predicate {\tt force ?o2 ?o2-force} which take value {\tt true} if {\tt ?o2-force} is a support surface. 

\begin{figure}[h!]
	\begin{center}
        \includegraphics[width=0.8\columnwidth]{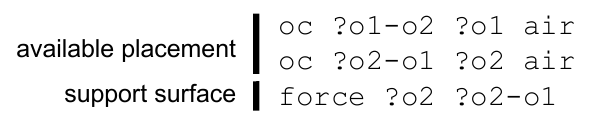}
		\caption{Predicates defining stable placement configurations for object-support pick-and-place tasks.
		}
		\label{fig:placement_constraint_pddl}
	\end{center}
\end{figure}

Fig. \ref{fig:placement_constraint_pddl} summarizes the predicates that evaluate placement constraints for placing actions in object-support pick-and-place tasks. Placement configurations {\tt ?o1-o2} and {\tt ?o2-o1} satisfying these constraints define the set of object-support poses $_2\xi_1$ so that {\tt StablePlace[o1,o2]}$(\:_2\xi_1) = 1$.

\subsubsection{Object-container Tasks}
\label{sec:object-container tasks}
We have seen in Sec. \ref{sec:object-support tasks} that object-support tasks focuses on single part interactions between the grasped object {\tt o1} and its support {\tt o2}, where object-support relations are fully characterized by the variables {\tt ?o1}, {\tt ?o2}, {\tt ?o1-o2} and {\tt ?o2-o1}. 
Let's assume now that we are dealing with tasks having constraints related to interactions of an object with multiple objects adjacent to it, not only with its support. These tasks are, for example, accommodating objects on a cluttered surface, where the pepper should be placed on a shelf, to the right of the salt, and in front of the thyme. 
To encode constraints in this type of tasks, we make use of object-container relations, where objects are placed inside container {\it spaces} (Fig. \ref{fig:example_place_types}D), rather than on support surfaces. In these {\it object-container} tasks, the manipulated object {\tt o1} is {\it absorbed} by its container space {\tt o2}, and we use the functional parts of {\tt o2} to encode constraints. 

The representation of TAMP constraints using predicates for object-container tasks is essentially similar to the representation for object-support tasks. As before, the main idea is to find the values for grasping, {\tt ?o1-h-p}, {\tt ?o1-h-f1}, and {\tt ?o1-h-f2} and placement configurations, {\tt ?o1-o2} and {\tt ?o2-o1}, that satisfy the grasping ({\tt StableGrasp}) and placement ({\tt StablePlace}) constraints described in Sec. \ref{sec:tamp_action_template}, respectively.
By using object-space relations we can track changes in multiple parts of the manipulated object {\tt o1} by only checking if this object is inside a container space {\tt o2} using the predicates {\tt in ?o2 ?o1}. With this representation, the number of variables to track changes is reduced to {\tt ?o1} and {\tt ?o2}, where the variables {\tt ?o1-o2} and {\tt ?o2-o1} take the constant value {\tt in}.

\begin{figure}[h!]
	\begin{center}
        \includegraphics[width=1\columnwidth]{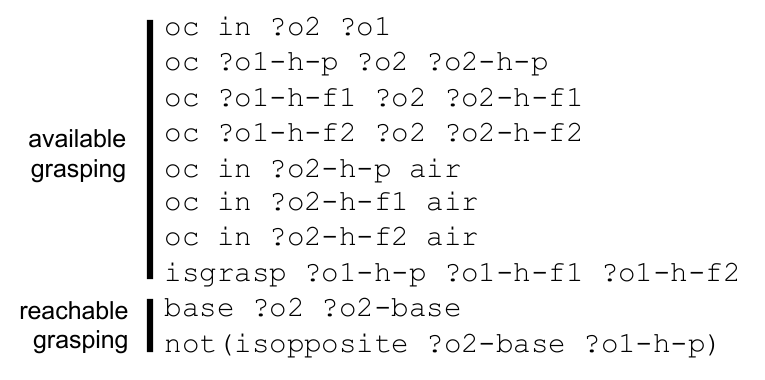}
		\caption{Predicates defining the grasping constraints for object-container pick-and-place tasks.
		}
		\label{fig:grasp_constraint_pddl_spaces}
	\end{center}
\end{figure}

In order to encode grasping constrains in object-container tasks, we evaluate hand-object relations in terms of the parts of the space {\tt o2} containing object {\tt o1}, rather than of the parts of {\tt o1}.
These constraints are considered by first checking if {\tt o1} is indeed contained by space {\tt ?o2} using the predicate {\tt in ?o2 ?o1}. Then, we identify the surrounding spaces that correspond to the evaluated grasping configuration using the predicates 
{\tt oc ?o1-h-p ?o2 ?o2-h-p}, {\tt oc ?o1-h-f1 ?o2 ?o2-h-f1}, and {\tt oc ?o1-h-f2 ?o2 ?o2-h-f2}, where {\tt ?o2-h-p}, {\tt ?o2-h-f1}, and {\tt ?o2-h-f2} are now spaces around the container space {\tt o2} interacting with the space parts {\tt ?o1-h-p}, {\tt ?o1-h-f1}, and {\tt ?o1-h-f2}, respectively. 
Then, we check whether these spaces are clear, which will allow a grasping {\tt o1}, using the predicates {\tt oc in ?o2-h-p air}, 
{\tt oc in ?o2-h-f1 air}, and {\tt oc in ?o2-h-f2 air}.
Kinematic feasibility is checked in a similar manner as in object-support tasks (Fig. \ref{fig:grasp_constraint_pddl}). However, in the object-container case, we use the side of the bounding box of space {\tt o2} to define the relative pose of the space with respect to the base of the robot, {\tt ?o2-base}.
Fig. \ref{fig:grasp_constraint_pddl_spaces} presents the predicates to check grasp constraints for picking actions in object-container tasks. Grasping configurations {\tt ?o1-h-p}, {\tt ?o1-h-f1}, and {\tt ?o1-h-f2} satisfying these constraints define the set of hand-object poses $_1\xi_h$ so that {\tt StableGrasp[o1,h]}$(\:_1\xi_h) = 1$.

\begin{figure}[h!]
	\begin{center}
        \includegraphics[width=0.75\columnwidth]{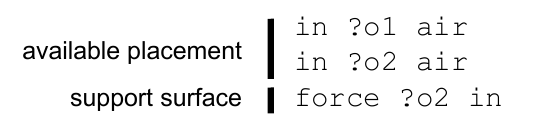}
		\caption{Predicates defining stable placement configurations for object-container pick-and-place tasks.
		}
		\label{fig:placement_constraint_pddl_spaces}
	\end{center}
\end{figure}

Placement constraints for object-container tasks are conceptually similar to those of object-support tasks (Fig. \ref{fig:placement_constraint_pddl}). In this case, variables defining the relations between the manipulated object {\tt o1} and its (container) support {\tt o2}, {\tt ?o2-o1} and {\tt ?o1-o2}, take the constant value {\tt in}, where available placements of an object in a space is evaluated by the predicate {\tt in ?o2 air}.
This predicate {\tt in ?o2 air} takes value {\tt true} if the space {\tt o2} is empty. 
In object-container tasks, spaces steadily hold objects when placed in it, which is reflected by the predicate {\tt force ?o2 in} taking value {\tt true}, where the variable {\tt ?o2-force} is replaced by the constant value {\tt in}.
Fig. \ref{fig:placement_constraint_pddl_spaces} shows the BOX predicates to evaluate placement constraint for placing actions in object-container pick-and-place tasks. Placement configurations satisfying these constraints define the set of object-support poses $_2\xi_1$ so that {\tt StablePlace[o1,o2]}$(\:_2\xi_1) = 1$.

\subsection{U-TAMP Actions}
\label{sec:U-TAMP Actions}
In Sec. \ref{sec:U-TAMP Constraints}, we have shown how the TAMP constraints are encoded using predicates and object-centric abstractions.
In this section, we include these constraints into U-TAMP action templates using PDDL notation compatible with task planning. 
This allows considering TAMP constraints in the heuristic search of task planning, yielding geometrically consistent plans that can be directly transformed into object and motion parameters for task execution, without any further sub-symbolic geometric reasoning. 
We define U-TAMP actions for object-support and object-container tasks, explaining in each case how these actions are related to the generic TAMP actions presented in Sec. \ref{sec:tamp_action_template}.
In the U-TAMP actions, we include predicates corresponding to the motion constraints for stable grasping ({\tt StableGrasp}) and placement ({\tt StablePlace}) (see Sec. \ref{sec:U-TAMP Constraints}). The kinematic constraint {\tt Kin} is calculated after task planning since the robot configuration $_r\xi_h$ is determined from the real-valued poses $_1\xi_h$ and $_2\xi_1$ obtained from the grounded actions of the task plan using the mechanisms described in Sec. \ref{sec:legal_kinematics}. 

\subsubsection{Object-support tasks}
\begin{figure}
	\begin{center}
		\subfigure[U-TAMP action {\tt Pick}]{
			\includegraphics[width=0.9\columnwidth]{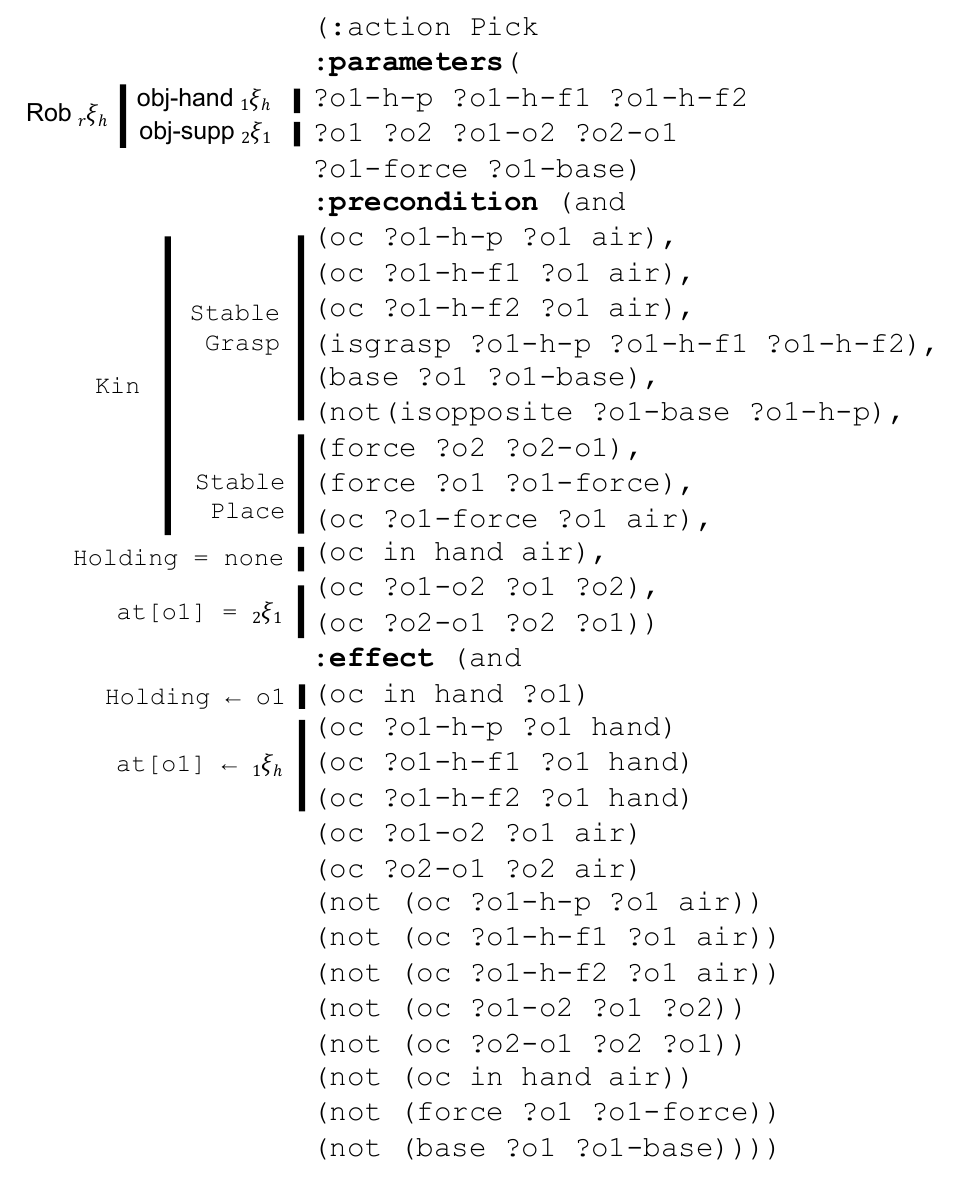}
			\label{fig:template_pick}
		}
		\subfigure[U-TAMP action {\tt Place}]{
			\includegraphics[width=1\columnwidth]{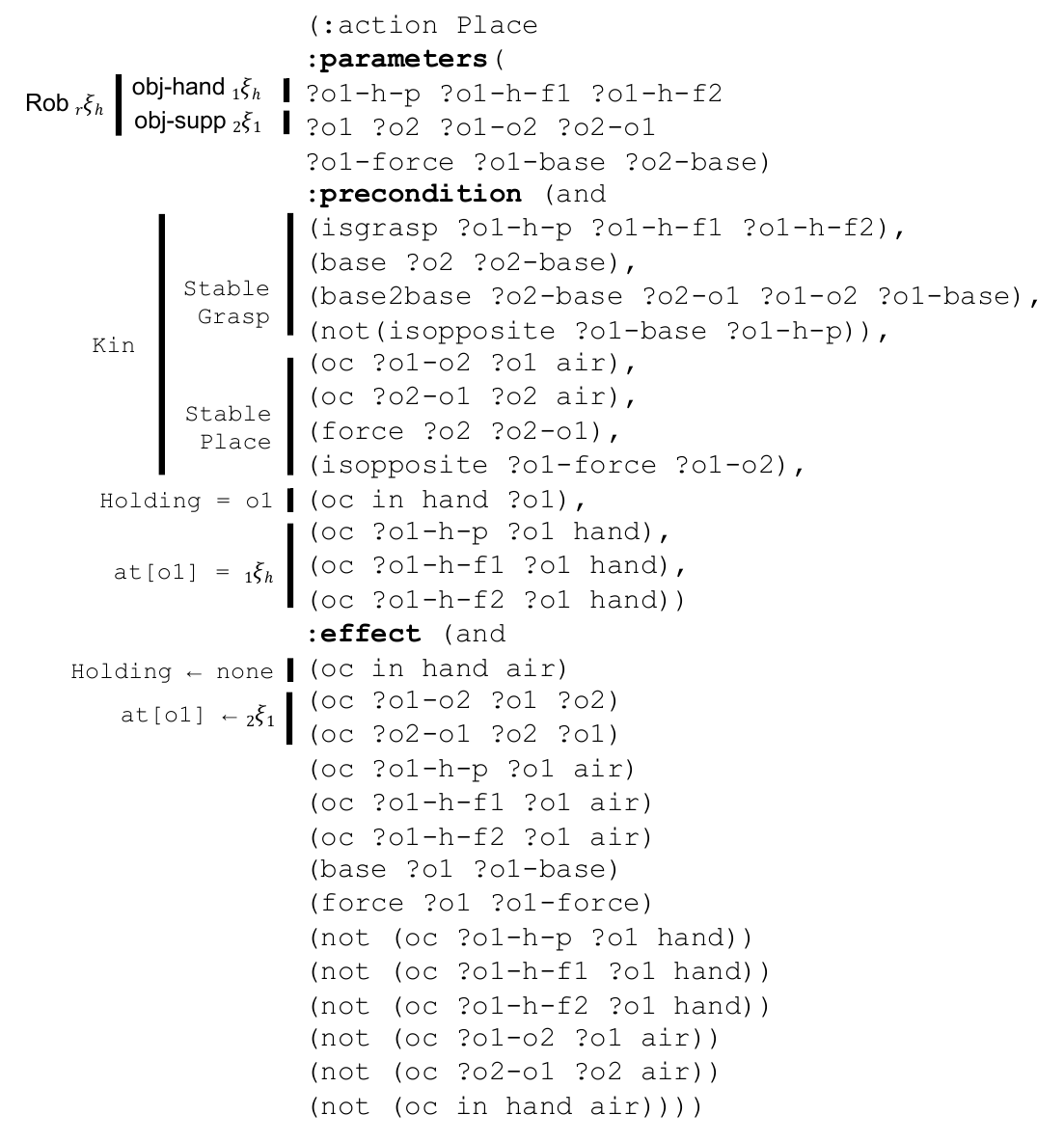}
			\label{fig:template_place}
		}
		\caption{U-TAMP action templates for object-support tasks.}
		\label{fig:template_pick_place}
	\end{center}
\end{figure}

Picking and placing actions in object-support tasks encode changes in the location of the manipulated object {\tt o1} from a support object {\tt o2} to the hand of the robot {\tt hand}, and vice-versa. In the {\tt grasp} and {\tt release} TAMP templates of Sec. \ref{sec:tamp_action_template}, these locations are encoded using the predicate {\tt at[o1]}, with value {\tt at[o1] = $_2\xi_1$} when the object is placed on its support and {\tt at[o1] = $_1\xi_h$} when the object is in the robot hand. 
Using the notation introduced in Sec. \ref{sec:U-TAMP Variables}, the object-support configuration {\tt at[o1] = $_2\xi_1$} can be encoded in terms of predicates as {\small \tt oc ?o1-o2 ?o1 ?o2} and {\small \tt oc ?o2-o1 ?o2 ?o1},
where the variables {\tt ?o1-o2} and {\tt ?o2-o1} represent the interacting parts of {\tt o1} and {\tt o2}. By assigning values to these variables, we can unequivocally calculate $_2\xi_1$, as explained in \ref{sec:legal_placement}. 
The object-hand configuration {\tt at[o1] = $_1\xi_h$}, in turn, can be encoded using predicates as {\small \tt oc ?o1-h-p ?o1 hand}, {\small \tt oc ?o1-h-f1 ?o1 hand}, and {\small \tt oc ?o1-h-f2 ?o1 hand},
where the variables {\tt ?o1-h-p}, {\tt ?o1-h-f1}, and {\tt ?o1-h-f2} represent the parts of {\tt o1} interacting with the parts of the robot hand. As before, by assigning values to these variables, we can unequivocally calculate $_1\xi_h$, as explained in \ref{sec:legal_grasping}. 
Fig. \ref{fig:template_pick_place} presents the {\tt Pick} and {\tt Place} task planning actions encoding changes in the object {\tt o1} location in PDDL notation for object-support tasks.
For completeness of the PDDL representation, we also consider the complementary changes by indicating the parts of object that become clear after the action in the effect of the PO in terms of the virtual object {\tt air}.
We also indicate in Fig.\ref{fig:template_pick_place} the analogies between the grounded predicates {\tt holding = none} and {\tt holding = o1} of the generic TAMP action templates (Fig. \ref{fig:tamp_templates}) with the predicates {\tt oc in hand air} and {\tt oc in hand ?o1}, respectively. 

The precondition part of the {\tt Pick} action includes predicates that evaluates available grasping configurations satisfying the {\tt StableGrasp} constraint of Fig. \ref{fig:grasp_constraint_pddl}.
The precondition part also includes predicates to evaluate placement constraints ({\tt StablePlace}) in a picking action. The predicate {\tt force ?o2 ?o2-o1} evaluates if the placement of {\tt o1} on {\tt o2} is stable. The predicates {\tt force ?o1 ?o1-force} and {\tt oc ?o1-force ?o1 air}, in turn, verify if {\tt o1} is not supporting any other object, which would make the object to fall down when {\tt o1} is picked. 
In the effect part of the {\tt Pick} action, the predicates {\tt force ?o1 ?o1-force} and {\tt base ?o1 ?o1-base} are deleted from the symbolic state, provided {\tt o1} will take arbitrary orientations during its manipulation after being picked. 

In the {\tt Place} action, the precondition part includes predicates that verify available placing configurations satisfying the placement constraints {\tt StablePlace} in Fig. \ref{fig:placement_constraint_pddl}. The precondition also includes predicates for evaluating if the current grasping is legal, {\tt isgrasp ?o1-h-p ?o1-h-f1 ?o1-h-f2}, and if the grasping configuration resulting from placing {\tt o1} on {\tt o2} will not compromise the robot kinematic. This latter assessment is similar to the one done in the {\tt Pick} action through the predicates {\tt base ?o1 ?o1-base} and {\tt not(isopposite ?o1-base ?o1-base ?o1-h-p)}. However, in the {\tt Place} case, the side of the bounding box of {\tt o1} closest to the robot base {\tt o1-base} is not available in the symbolic state since it was deleted in a previous picking action. Fortunately, it can be derived from the object-support configuration resulting from placing {\tt o1} on {\tt o2}. To this end, we define the predicate {\tt base2base ?o2-base ?o2-o1 ?o1-o2 ?o1-base} that maps the part of the support object {\tt o2} closest to the base, {\tt ?o2-base}, present in the symbolic state, to the part of {\tt o1} that will be closest to the base, {\tt ?o1-base}, after the placing action. 

After placing {\tt o1} on a surface of {\tt o2}, we need to restore the predicates necessary for future evaluations of the {\tt StableGrasp} and {\tt StablePlace} constraints, i.e. to make {\tt o1} pick-able and being able to support other objects for future placing and placing actions, respectively. 
The {\tt StableGrasp} constraint needs the predicate {\tt base ?o1 ?o1-base}, which is added to the symbolic state in the effect part of the action {\tt Place}. On the other hand, in order to define which part of {\tt o1} can be used to support other objects in future placing actions ({\tt StablePlace} constraint), we {\it propagate} the force of {\tt o2} into {\tt o1} through the predicate {\tt isopposite ?o1-force ?o1-o2} in the precondition part and add the predicate {\tt force ?o1 ?o1-force} in the effect part of action {\tt Place}.

\subsubsection{Object-container tasks}
\begin{figure}
	\begin{center}
		\subfigure[U-TAMP action {\tt Pick-space}]{
			\includegraphics[width=1\columnwidth]{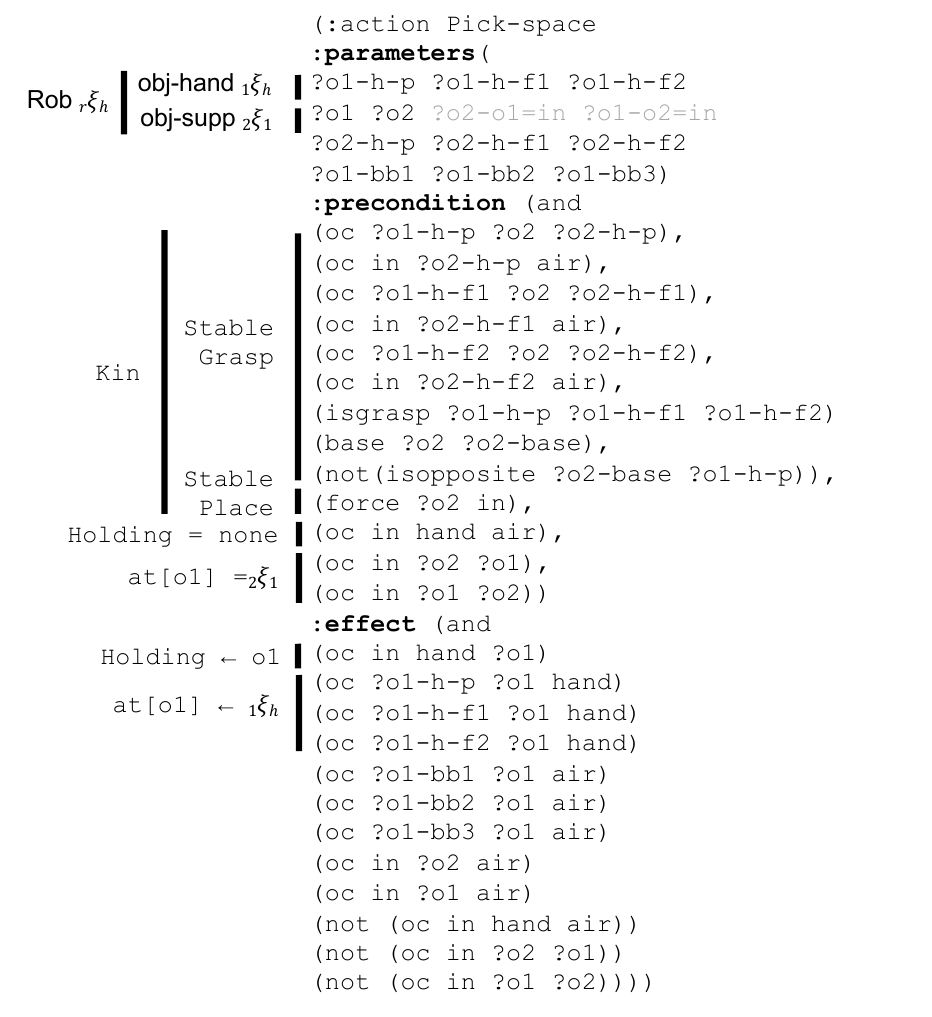}
			\label{fig:example_cup}
		}
		\subfigure[U-TAMP action {\tt Place-space}]{
			\includegraphics[width=1\columnwidth]{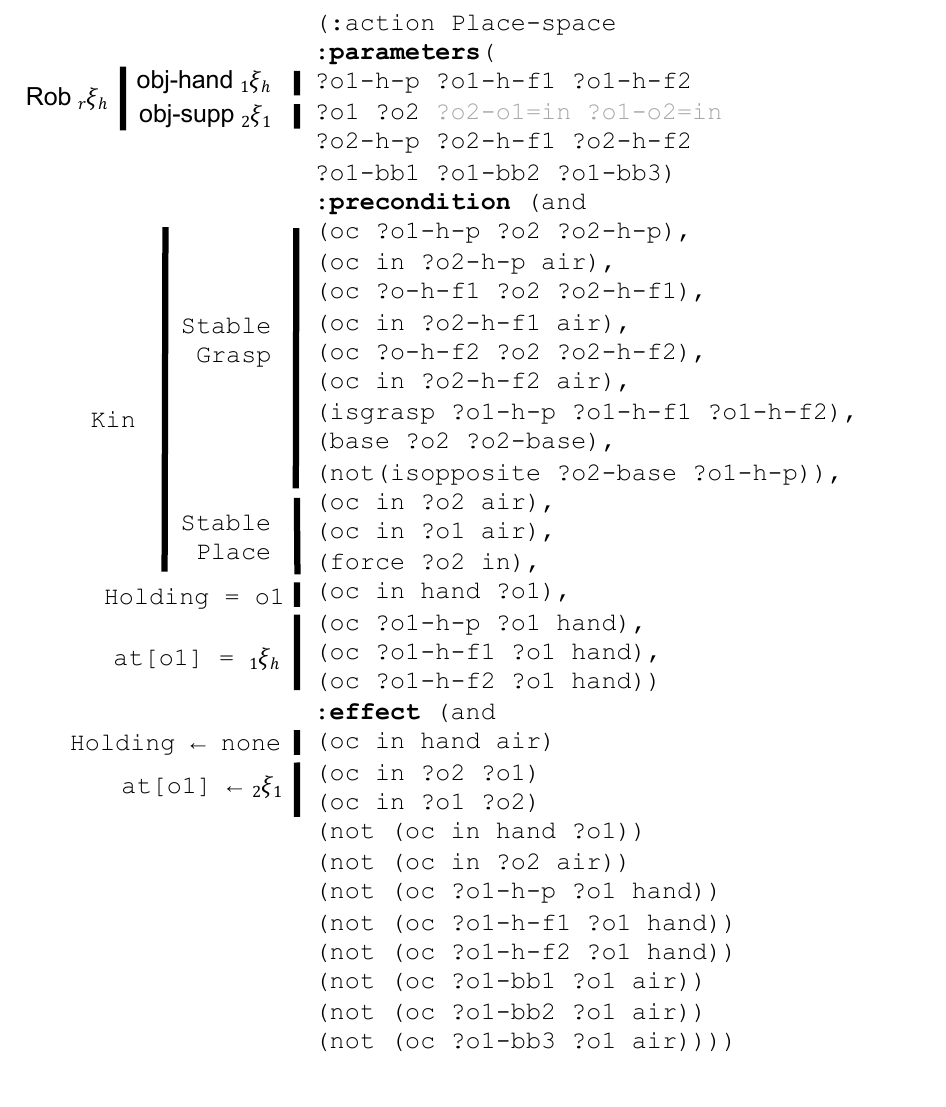}
			\label{fig:example_AB}
		}
		\caption{U-TAMP action templates for object-container tasks.}
		\label{fig:template_pick_place_spaces}
	\end{center}
\end{figure}

In object-container task, picking and placing actions are named as {\tt Pick-space} and {\tt Place-space}, respectively, to distinguish them from the {\tt Pick} and {\tt Place} actions for object-support tasks. 
These actions encode changes in the location of the manipulated object {\tt o1}, now from a {\it container} space {\tt o2} to the hand of the robot {\tt hand}, and vice-versa. This representation is done using the same notation as in the object-support task with the only difference that the object-support relations {\tt ?o1-o2} and {\tt ?o2-o1} takes the constant value {\tt in}. Fig. \ref{fig:template_pick_place_spaces} shows the {\tt Pick-space} and {\tt Place-space} actions.

The precondition part of the {\tt Pick-space} action contains the predicates that evaluates available grasping configurations satisfying the constraint {\tt StableGrasp}, where the grasp availability is evaluated by checking if the spaces adjacent to the container space {\tt o2}, {\tt o2-h-p}, {\tt o2-h-f1}, and {\tt o2-h-f2}, are empty, rather than if the sides of the contained object {\tt o1} are clear, as explained in Sec. \ref{sec:object-container tasks} and illustrated in Fig. \ref{fig:grasp_constraint_pddl_spaces}. 
The precondition part also includes the predicate {\tt force ?o2 in} to evaluate if the placement of {\tt o1} inside {\tt o2} is stable. 
Since the {\tt force} and {\tt base} predicates are evaluated for the space {\tt o2} rather than for the manipulated object {\tt o1} and provided the location of spaces are fixed throughout the task in object-container tasks, it is not necessary to delete or add them back after a picking or placing action. 

{\tt Pick-space} and {\tt Place-space} action templates are defined considering that they will be part of domain definitions in {\it hybrid} object-support and object-container tasks. Thus, after an object is picked from a space, we should make it available for either placing it inside another space or placing it on a support. To this end, we need to indicate which sides of its bounding box become available for future placing-on-support actions. This is done by adding in the effect part of the {\tt Pick-space} action the predicates {\tt ?o1-bb1 ?o1 air}, {\tt ?o1-bb2 ?o1 air}, and {\tt ?o1-bb3 ?o1 air}, where {\tt ?o1-bb1}, {\tt ?o1-bb2}, and {\tt ?o1-bb3} represents the parts of {\tt o1} that are not interacting with the hand. 
For consistency of the PDDL representation, we need to explicitly check which parts of {\tt o1} are not interacting with the hand. This is done by including in the precondition part predicates of the form {\tt not(isequal ?o1-bb ?o1-h)}, where {\tt ?o1-bb} takes values {\tt ?o1-bb1}, {\tt ?o1-bb2}, and {\tt ?o1-bb3} and {\tt ?o1-h} takes values {\tt ?o1-h-p}, {\tt ?o1-h-f1}, and {\tt ?o1-h-f2}. These predicates were not included in the figure for simplicity.

The precondition part of the {\tt Place-space} includes predicates that verify available placing configurations satisfying the placement constraint {\tt StablePlace} for object-container tasks (Fig. \ref{fig:placement_constraint_pddl_spaces}. In addition, the precondition part includes the same {\tt StableGrasp} constraints as  the {\tt Pick-space} action. These constraints permit checking if there is enough room for the hand around the space {\tt o2} to place {\tt o1} inside it without colliding with adjacent objects. 
Since the container space {\tt o2} {\it absorbs} {\tt o1} after the placing action, we need to delete from the symbolic state those predicates describing the relations of {\tt o1} with its surrounding to keep the geometrical consistency in the planning process. This is done by deleting all the predicates representing such relations in the effect part of {\tt Place-spaces}, e.g. {\tt not(oc ?o1-bb1 ?o1 air)}.

\subsection{U-TAMP Perception}
\label{sec:U-TAMP Perception}
In this section, we define the functions to map {\tt true} (or {\tt false}) values to predicates represented using object-centric abstractions of constraints from the poses and bounding boxes of the objects in the scenario, obtained from sensing mechanisms. 

\subsubsection{Object-object interactions}
For the evaluation of predicates defining object-centric interactions between adjacent objects {\tt oc ?o1-o2 ?o1 ?o2} (e.g. object-support, object-container, and object-hand relations), we resort again to the use of object-centric abstractions and define a space {\tt s-o1-o2} associated to the part {\tt o1-o2} of object {\tt o1}. 
This space has size $\Delta_{\tt s-o1-o2} = \Delta_{\tt 1}$
and pose $_r\xi_{\tt s-o1-o2} = \{\:_rp_{\tt s-o1-o2}, \:_rw_{\tt s-o1-o2}\}$, where {\tt s-o1-o2} has the same orientation as object {\tt o1}, $\:_rw_{\tt s-o1-o2} = \:_rw_1$, and $_rp_{\tt s-o1-o2} =\:_rp_1 + \:_rR_1 \:_1p_{\tt s-o1-o2}$, where $_1p_{\tt s-o1-o2}$ is the position of space {\tt s-o1-o2} in the reference frame of object {\tt o1} (see Table \ref{table:ops}). 
We say that a grounded predicate {\tt oc o1-o2 o1 o2} takes value {\tt true} if $_rp_2 \in$ {\tt s-o1-o2}, where $_rp_2$ is the position of the centroid of object {\tt o2} in the global reference frame and {\tt s-o1-o2} is the space associated to the part {\tt o1-o2} of {\tt o1}. In the same line, we say that the grounded predicate {\tt oc o1-o2 o1 air} takes value {\tt true} if none of the centroids of the objects in the scenario is inside {\tt s-o1-o2}.
\begin{table}[h!]
\caption{Position $_1p_{\tt s-o1-o2}$ of space {\tt s-o1-o2} in the reference frame of object {\tt o1} for different values of {\tt o1-o2}}
{\small
\begin{center}
\resizebox{0.8\columnwidth}{!}{
\begin{tabular}{llll}
{\tt o1-o2} & $_1x_{\tt s-o1-o2}$ & $_1y_{\tt s-o1-o2}$ & $_1z_{\tt s-o1-o2}$ \\
\hline
{\tt on} & 0 & 0 & $\Delta^z_1$ \\
{\tt under} & 0 & 0 & $-\Delta^z_1$\\
{\tt left} & 0 & $\Delta^y_1$ & 0 \\
{\tt right} & 0 & $-\Delta^y_1$ & 0 \\
{\tt front} & $\Delta^x_1$ & 0 & 0 \\
{\tt back} & $-\Delta^x_1$ & 0 & 0 \\
{\tt in} & 0 & 0 & 0 \\
\end{tabular}}
\end{center}}
\label{table:ops}
\end{table}

\subsubsection{Object-base relations}
To determine the surface {\tt o1-base} for which the predicate {\tt base o1 o1-base} takes value {\tt true} we select the surface {\tt o1-base} of object {\tt o1} whose centroid, calculated as in Table \ref{table:oph}, is the closest to the centroid of the base of the robot, i.e. {\tt o1-base}$= \argmin_{\tt o1-side} ||\:_rp_{\tt o1-side} - \:_rp_{\tt base}||$. 

\subsubsection{Object-force relations}
The support surface {\tt o1-force} of an object {\tt o1} for a positive evaluation of the predicate {\tt force o1 o1-force} is determined using the normal of the surface {\tt o1-force} and the normal of the reference support surface (e.g. table top). If these normal vectors are parallel and in the same direction, we say that {\tt o1-force} is a support surface and the predicate {\tt force o1 o1-force} takes value {\tt true}.
\subsection{U-TAMP Execution}
\label{sec:U-TAMP Execution}
This section provides the basic mechanisms to transform a U-TAMP plan into motion parameters for plan execution. 
%
U-TAMP {\tt Pick} and {\tt Pick-space} actions in 
 a task plan are decomposed into {\tt moveF}$(\: _r\xi_h^{ini},\: \tau, \:_r\xi_h^{end})$ and {\tt grasp[o1 o2]}$(\:_r\xi_h, \:_2\xi_1, \:_1\xi_h)$ TAMP actions (Sec. \ref{sec:tamp_action_template}) using the symbol-to-pose transformations explained in Sec. \ref{sec:U-TAMP Variables}. 
The hand-object pose $_1\xi_h$ for the {\tt grasp} action is obtained from the grounded variables {\tt o1-h-p}, {\tt o1-h-f1}, and {\tt o1-h-f2}, as explained in Sec. \ref{sec:legal_grasping}. In the same line, the object-location pose $_2\xi_1$ is obtained from the grounded variables {\tt o1-o2} and {\tt o2-o1} using the mechanisms described in Sec. \ref{sec:legal_placement}. The final pose of the hand $\:_r\xi_h^{end}$ for grasping object {\tt o1} is obtained from $_1\xi_h$ and $_2\xi_1$ using Eqs. \eqref{eq:_rp_h}-\eqref{eq:_rR_1} (Sec. \ref{sec:legal_kinematics}).
Note that the values $_1\xi_h$, $_2\xi_1$, and $_r\xi_h^{end}$ already fulfill the constraints {\tt StableGrasp}, {\tt StablePlace}, and {\tt Kin}, provided these constraints are considered in the U-TAMP actions (Sec. \ref{sec:U-TAMP Actions}).

For the generation of a motion trajectory $\tau$ that fulfills the constraints {\tt CFree}$(\tau)$ and {\tt Motion}$(\: _r\xi_h^{ini},\: \tau, \:_r\xi_h^{end})$ for the execution of the {\tt moveF} action we define a pre-grasping point $_r\xi_h^{pg}$ and then generate a 6D motion trajectory using a cubic spline interpolation connecting the points $(\:_r\xi_h^{pg}, \:_r\xi_h^{pg}, \:_r\xi_h^{pg})$.
The pre-grasping pose $_r\xi_h^{pg}$ is calculated using an object-centric approach as
\begin{equation}
    _rp^{pg}_h = \:_rp_1 + \:_rR_1 \:_1p^{pg}_h,
\end{equation}
\noindent where $_1p^{pg}_h$ is the final pre-grasp position of the hand in the reference frame of the to-be-grasped object {\tt o1}. The values of $_1p^{pg}_h$ are obtained from multiplying the relative hand-object positions $_1p_h$ of Table \ref{table:oph} by a factor of three so as to approach the object from a safe distance before grasping. The pre-grasping orientation of the hand is defined as $_rR^{pg}_h = \:_rR_h$, i.e. as the same orientation of the final grasping pose. 
We would like to point out that this is a simple approach that works well for the pick-and-place tasks considered in this work (Sec. \ref{sec:experiments}). However, for other type of tasks, a more elaborated collision-avoidance mechanisms might be needed.

Similarly to picking actions, U-TAMP {\tt Place} and {\tt Place-space} in a task plan are decomposed into {\tt moveH}$(\: _r\xi_h^{ini},\: \tau, \:_r\xi_h^{end})$ and {\tt grasp[o1 o2]}$(\:_r\xi_h, \:_2\xi_1, \:_1\xi_h)$ TAMP actions using the symbol-to-pose transformations explained in Sec. \ref{sec:U-TAMP Variables}. 
In the same vein of the grasping case, we define a {\it pre-placing} pose $_r\xi^{pp}_h = \{\:_rp^{pp}_h, \:_rR^{pp}_h \}$ for the execution of the {\tt moveH} action. The position $_rp^{pp}_h$ is calculated as
\begin{equation}
    _rp^{pp}_h = \:_rp^{pp}_1 + \:_rR_1 \:_1p_h,
\end{equation}
\noindent where
\begin{equation}
    _rp^{pp}_1 =\:_rp_2 + \:_rR_2 \:_2p^{pp}_1
\end{equation}
\noindent is calculated using the sensed pose of the support object $_r\xi_2 = \{\:_rp_2, \:_rR_2 \}$ and $_2p^{pp}_1$, this latter obtained from multiplying the values in Table \ref{table:ops} by a factor of three so as to approach the final placement pose from a safe distance. The pre-placing orientation is defined as $_rR^{pp}_h = \:_rR_h$.

\section{EXPERIMENTS}
\label{sec:experiments}
The validity of our U-TAMP framework was assessed in two challenging non-monotonic tasks of arranging green blocks on a cluttered surface and of cooking dinner for two. These tasks are variations of the one proposed by Garrett et al. \cite{garrett2018ffrob}.
For all the experiments, we use the off-the-shelf planner Fast Downward \cite{helmert2006fast} with a lazy greedy best-first search with preferred operators.
\begin{figure}[h!]
	\begin{center}
		\subfigure[Initial State]{
			\includegraphics[width=0.46\columnwidth]{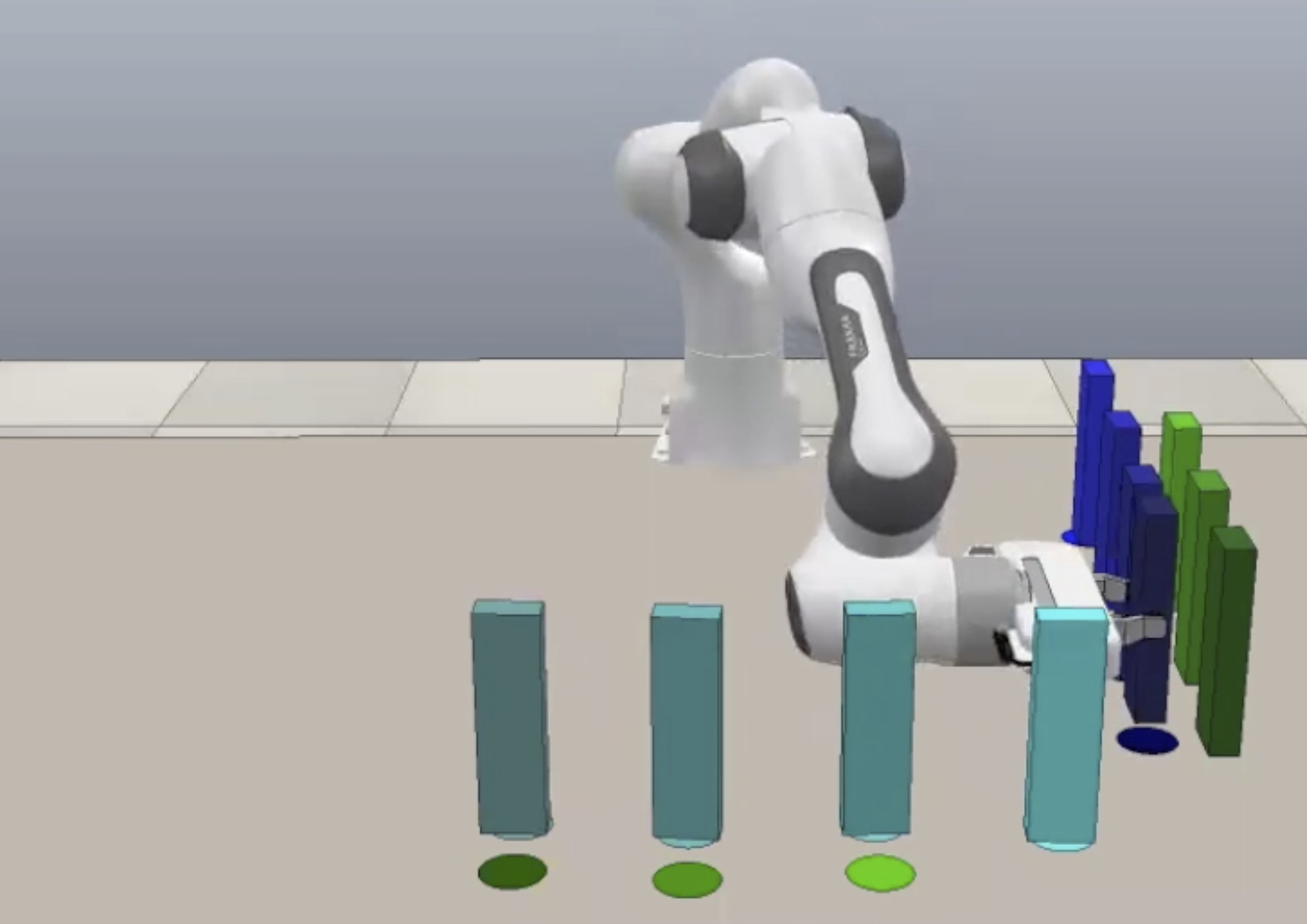}
			\label{fig:experiments_problem_32_ini}
		}
		\subfigure[Goal State]{
			\includegraphics[width=0.46\columnwidth]{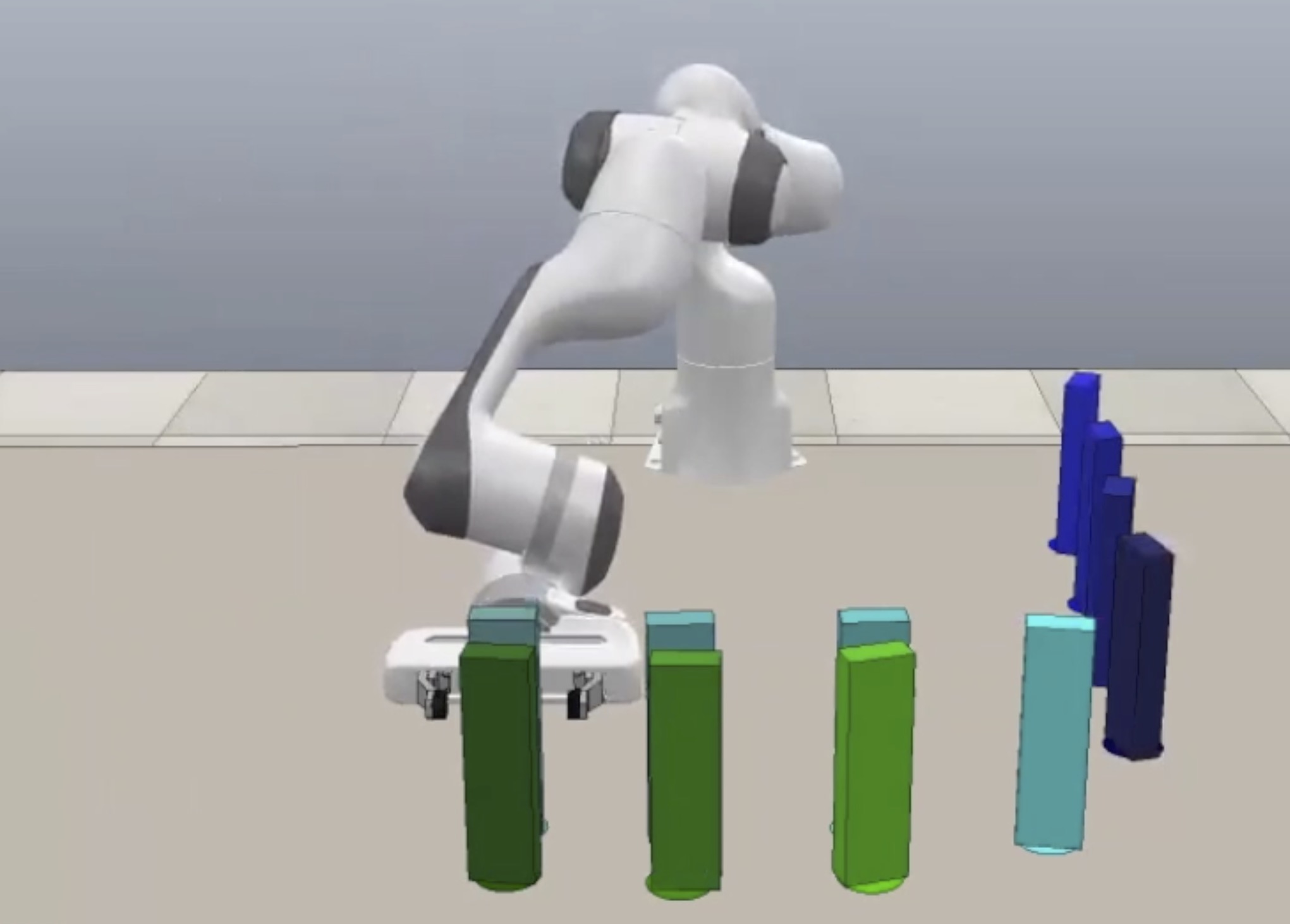}
			\label{fig:experiments_problem_32_goal}
		}
		\caption{Snapshots of example initial and goal states for the object-container task of arranging green blocks, adapted from \cite{garrett2018ffrob}.}
		\label{fig:experiments_problem_32}
	\end{center}
\end{figure}
\begin{figure}[h!]
	\begin{center}
		\subfigure[Initial State]{
			\includegraphics[width=0.46\columnwidth]{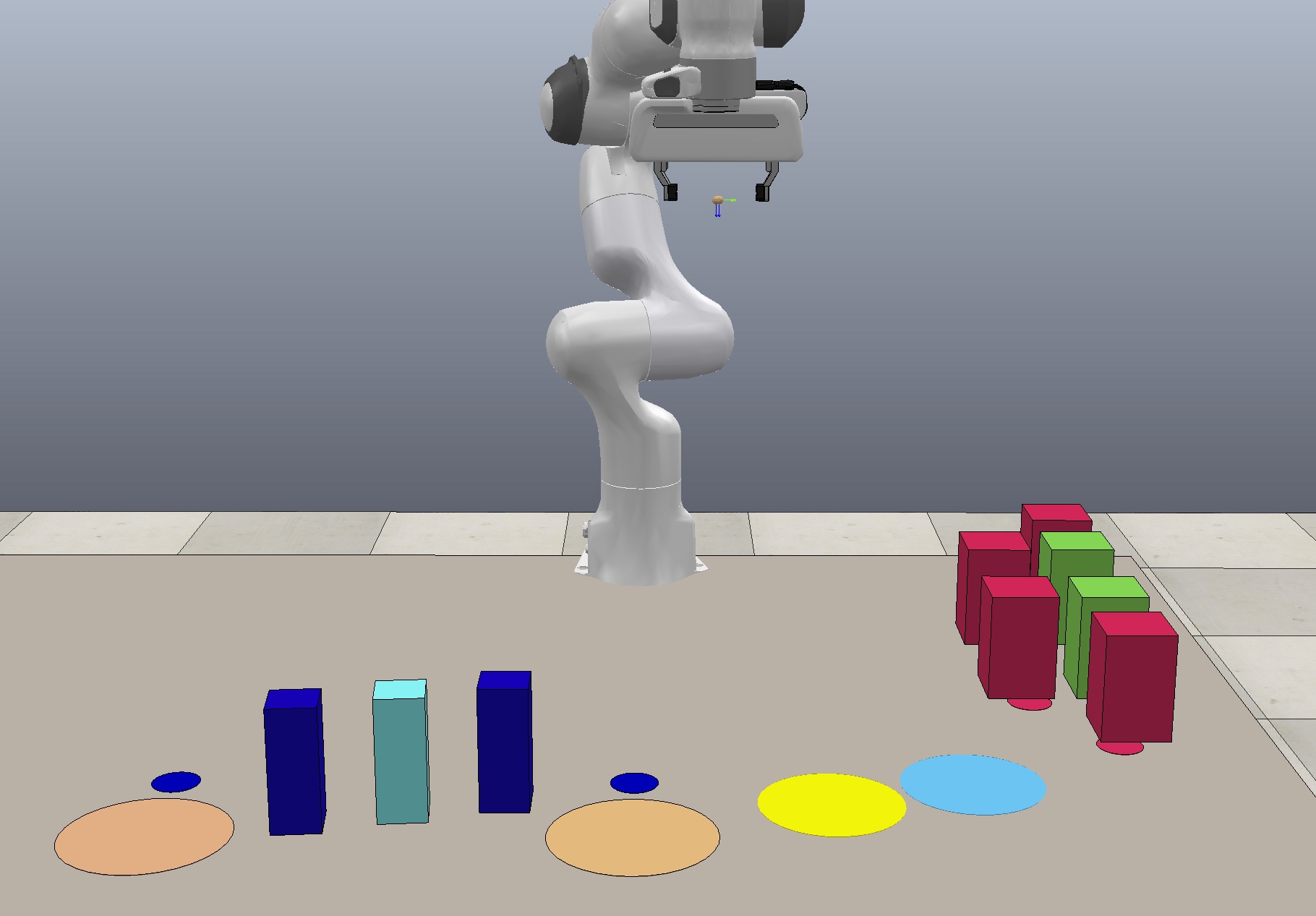}
			\label{fig:experiments_problem_5_ini}
		}
		\subfigure[Goal State]{
			\includegraphics[width=0.46\columnwidth]{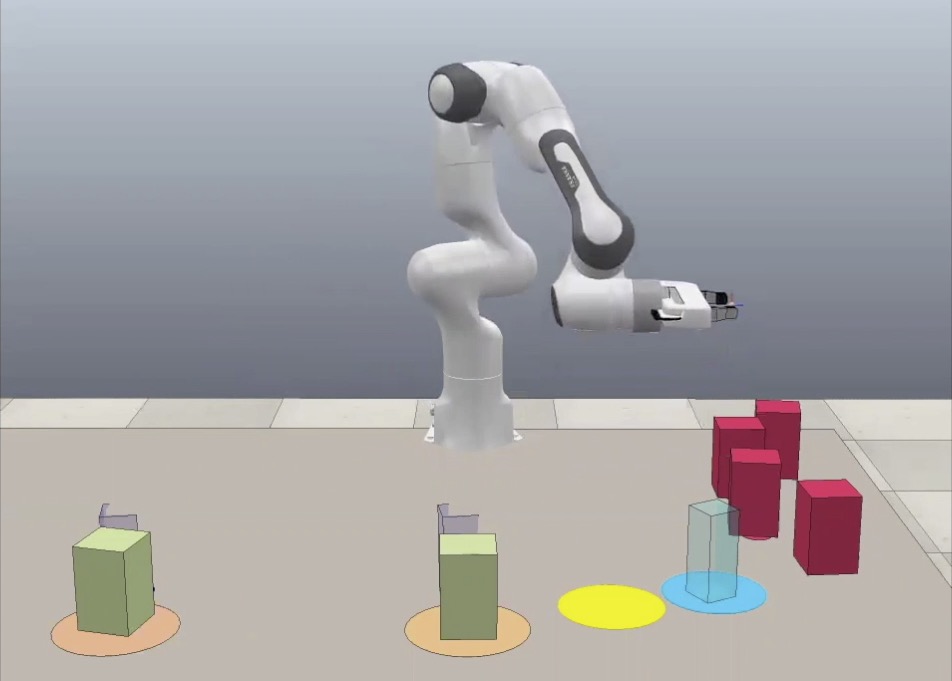}
			\label{fig:experiments_problem_5_goal}
		}
		\caption{Snapshots of example initial and goal states for the hybrid object-support and container task of cooking dinner for two, adapted from \cite{garrett2018ffrob}.}
		\label{fig:experiments_problem_5}
	\end{center}
\end{figure}

\subsection{Task 1: Arranging Green Blocks}
The scenario for this task is depicted in Fig. \ref{fig:experiments_problem_32}. This task (Problem 32 in \cite{garrett2018ffrob}) consists in moving the green blocks behind the blue blocks to their corresponding positions behind the cyan blocks. Blocks cannot be grasped from top or back, and blue and cyan blocks should be placed back in their initial positions. 
This makes the problem highly non-monotonic since the robot has to repeatedly undo goal conditions in order to complete the task. 
To solve this task, we adopt the object-container approach (Sec. \ref{sec:U-TAMP Constraints} and \ref{sec:U-TAMP Actions}) since the task requires satisfying constraints involving simultaneous interactions with multiple parts of objects. Hence, we use the planning operators {\tt Pick-space} and {\tt Place-space} (Fig. \ref{fig:template_pick_place_spaces}) for the planning domain definition. 
Since blocks cannot be grasped from top or back, we restrict the valid grasping configurations to  {\tt isgrasp(front,left,right)}, which takes value {\tt true} when {\tt ?o1-h-p = front}, {\tt ?o1-h-f1 = left}, and {\tt ?o1-h-f2 = right}.

The problem definition comprises a set of blue blocks, {\tt bblue}$i$, $i=1,..,4$, a set of cyan blocks, {\tt bcyan}$j$, $j=1,..,4$, and a set of green blocks, {\tt bgreen}$k$, $k=1,..,3$. These blocks are originally placed in their corresponding spaces {\tt sblue}$i$, {\tt scyan}$j$, and {\tt sgreen}$k$.
We also include two additional spaces on the table that can be used to temporally place blocks, {\tt stable1} and {\tt stable2}. Target spaces for the green blocks are labelled as {\tt sgreeng}$k$. Adjacent spaces to assess grasping constraints are also considered.

\begin{figure}
	\begin{center}
		\subfigure[{\tt Pk-s(bblue4,sblue4)}]{
			\includegraphics[width=0.46\columnwidth]{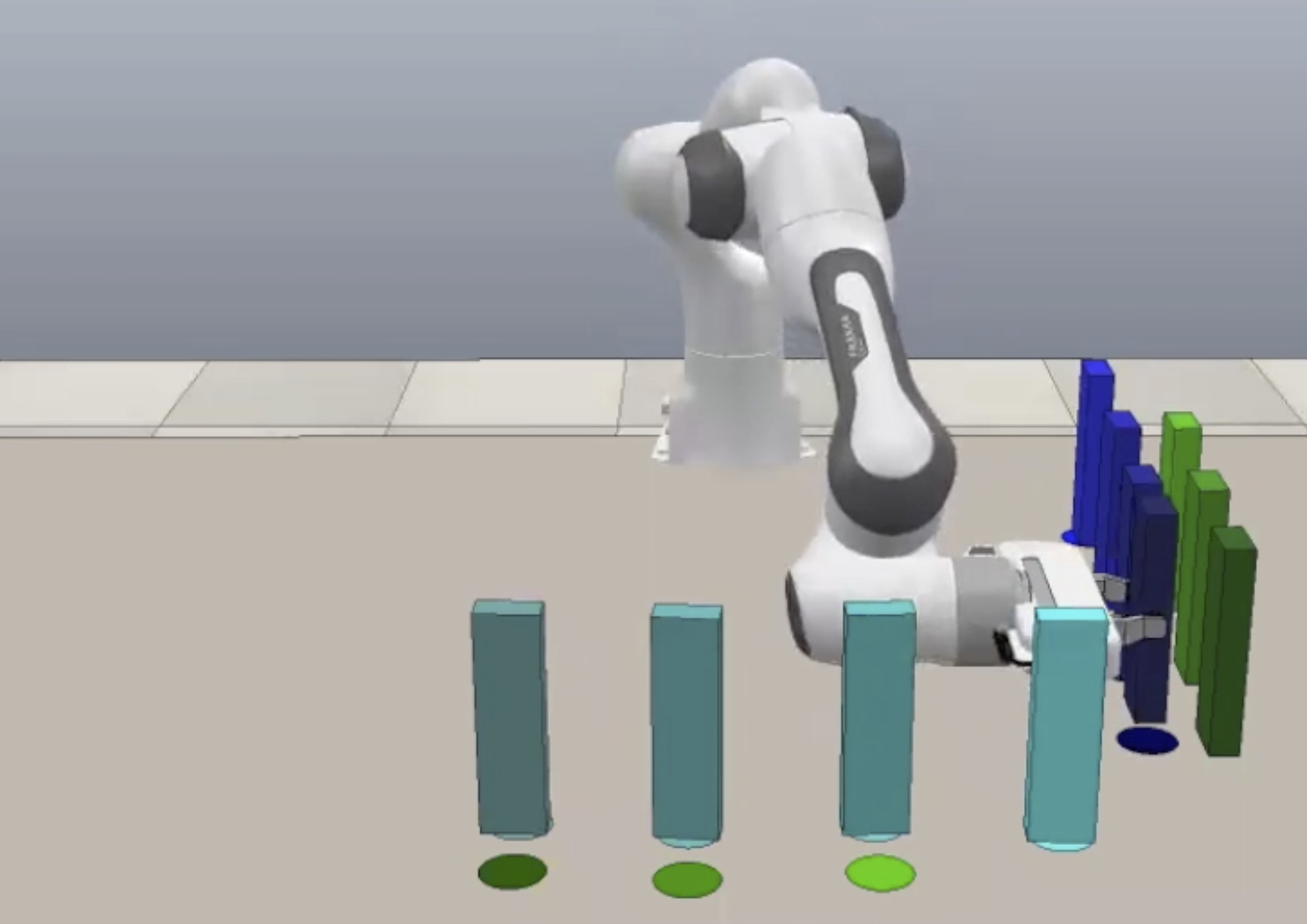}
			\label{fig:p32_1}
		}
		\subfigure[{\tt Pl-s(bblue4,stable1)}]{
			\includegraphics[width=0.46\columnwidth]{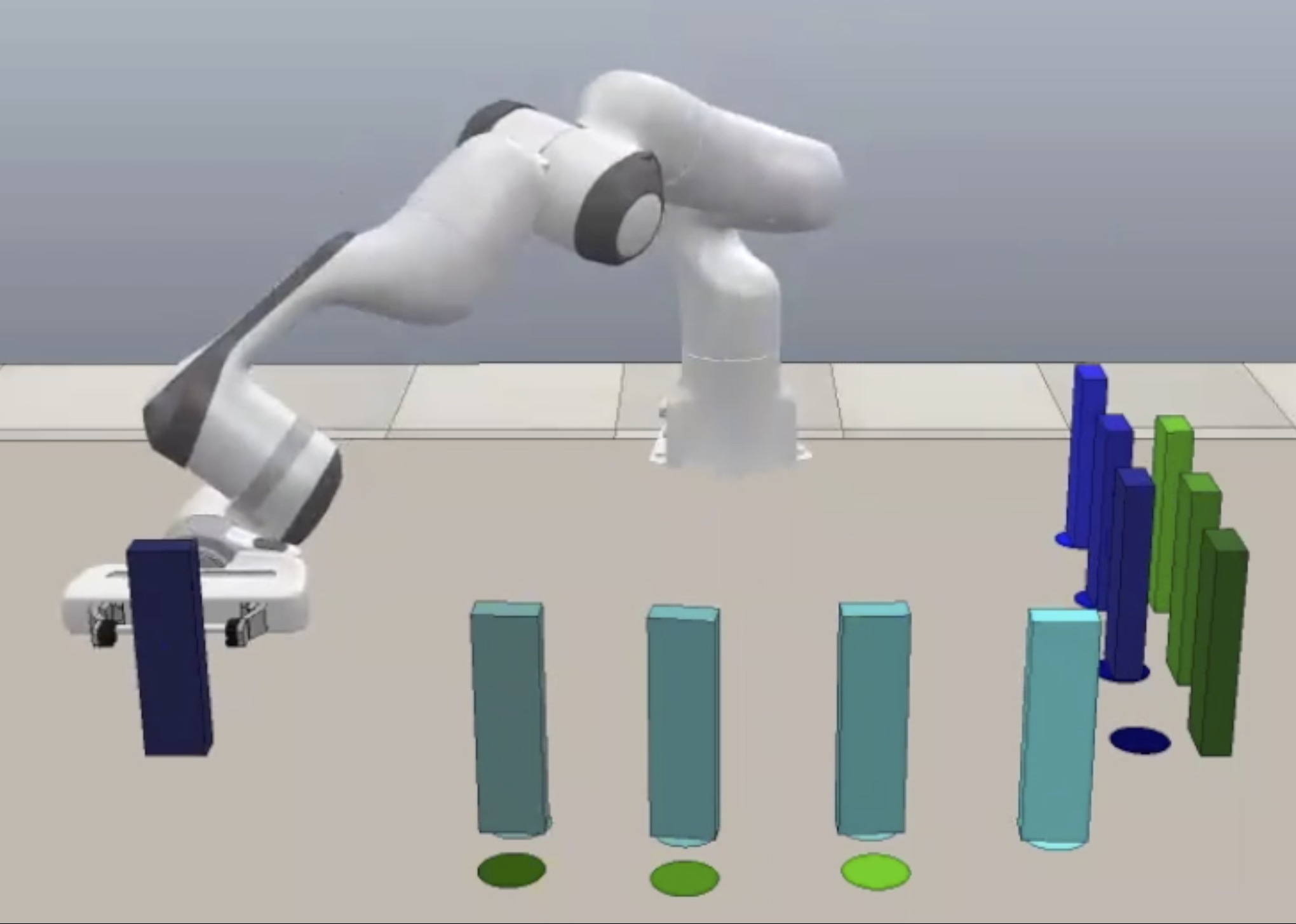}
			\label{fig:p32_2}
		}
        \subfigure[{\tt Pk-s(bgreen3,stable2)}]{
			\includegraphics[width=0.46\columnwidth]{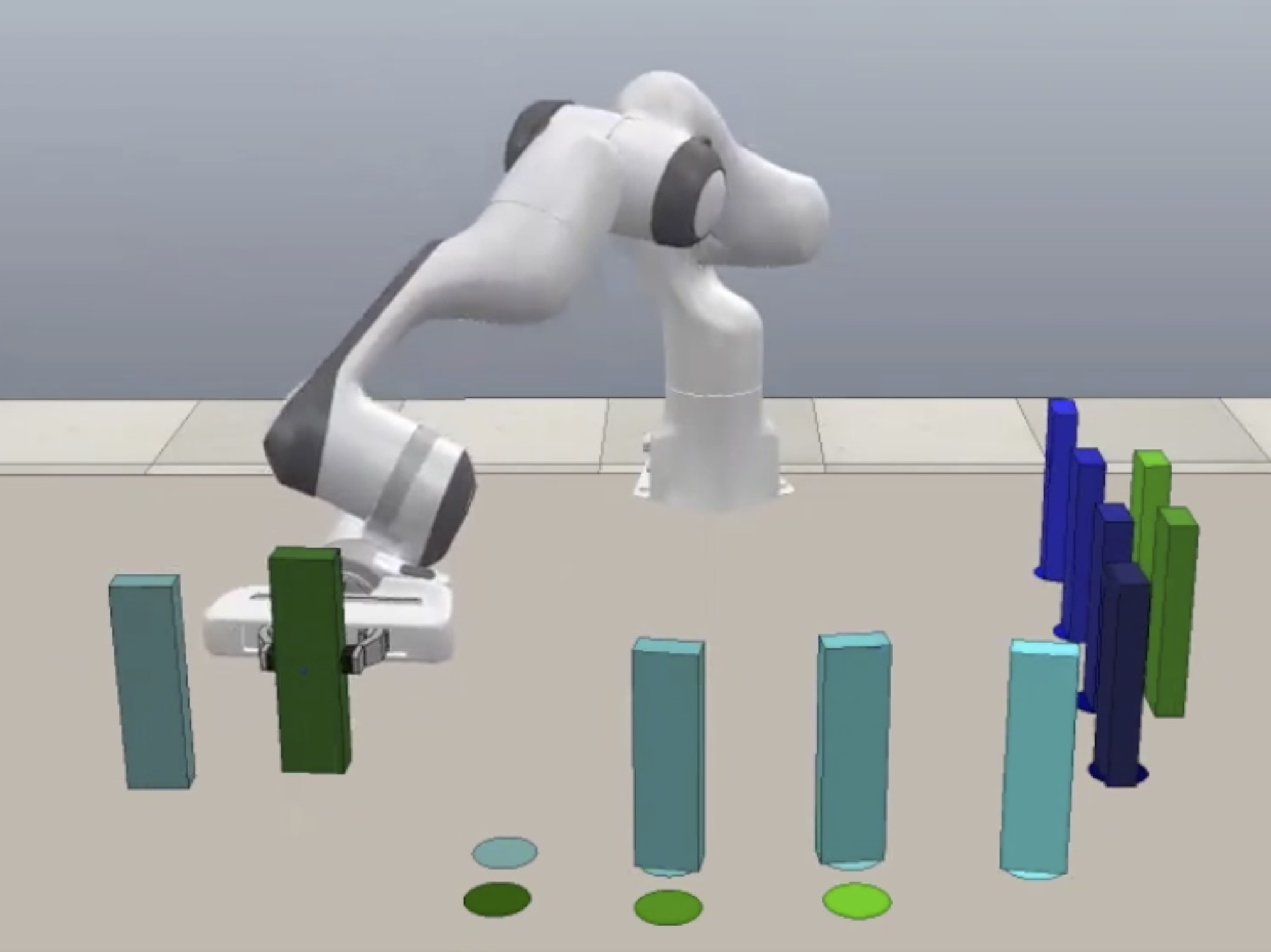}
			\label{fig:p32_3}
		}
		\subfigure[{\tt Pl-s(bgreen3,sgreeng3)}]{
			\includegraphics[width=0.46\columnwidth]{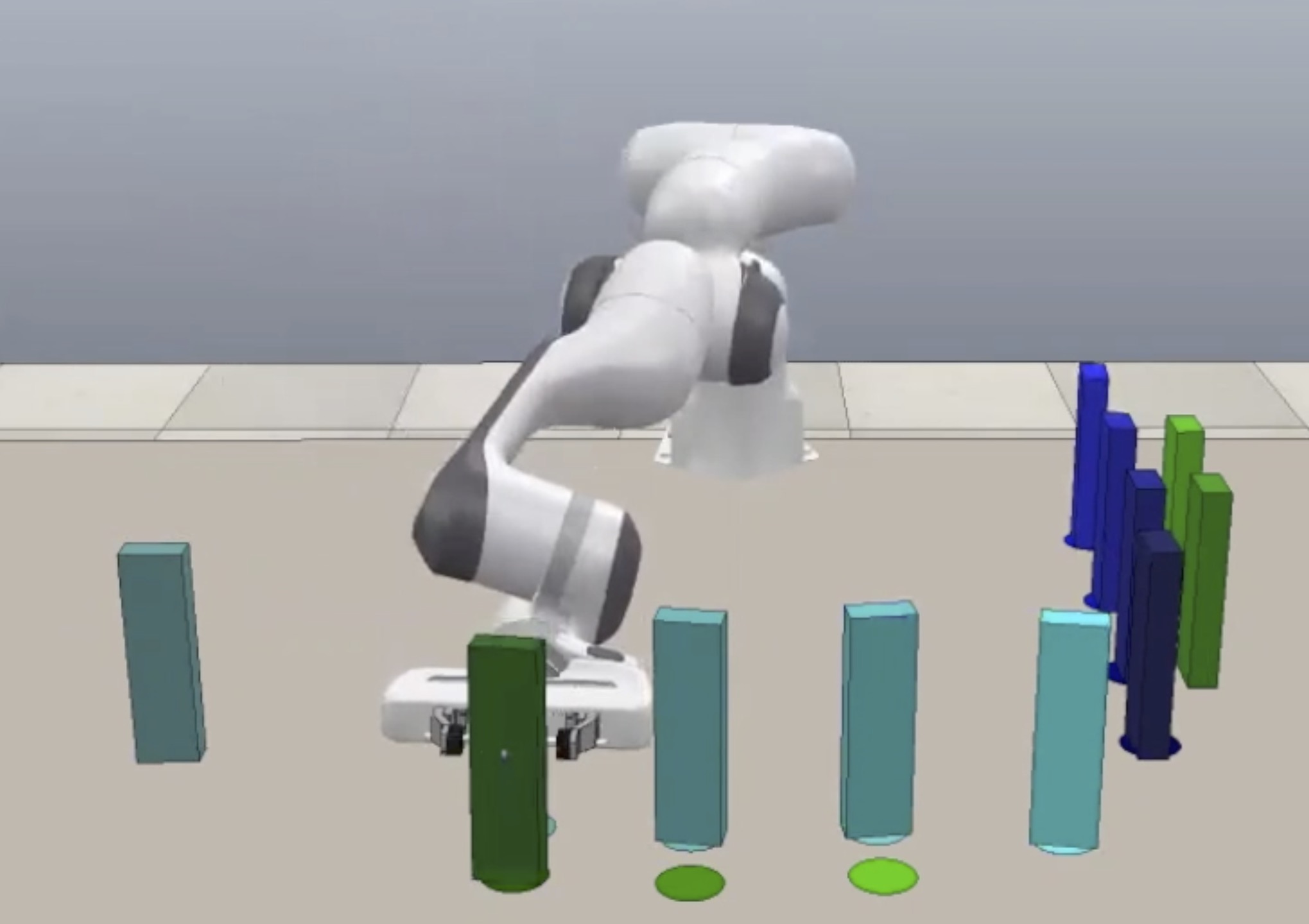}
			\label{fig:p32_4}
		}
        \subfigure[{\tt Pk-s(bgreen2,sblue3)}]{
			\includegraphics[width=0.46\columnwidth]{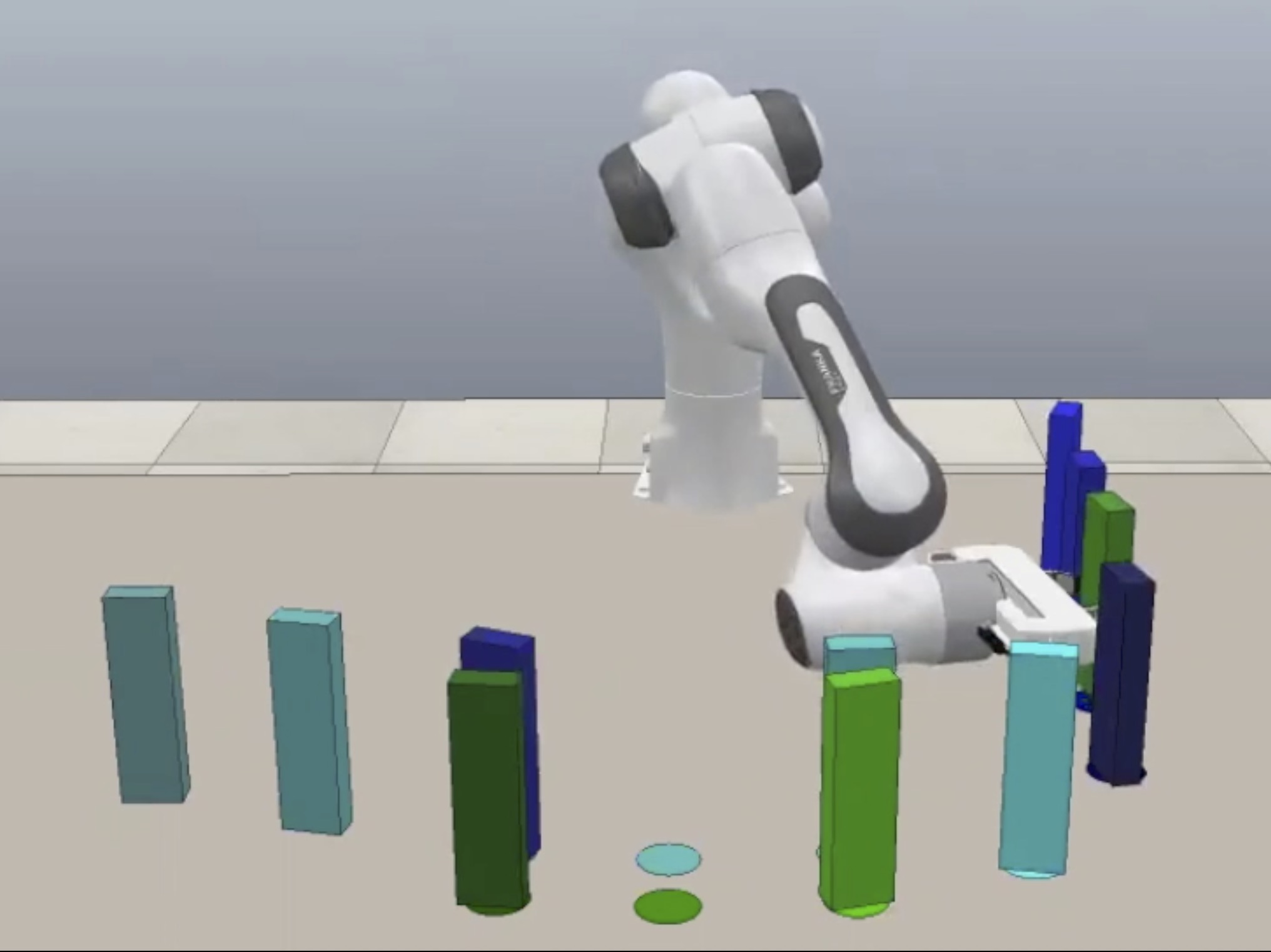}
			\label{fig:p32_5}
		}
		\subfigure[{\tt Pl-s(bgreen2,sgreeng2)}]{
			\includegraphics[width=0.46\columnwidth]{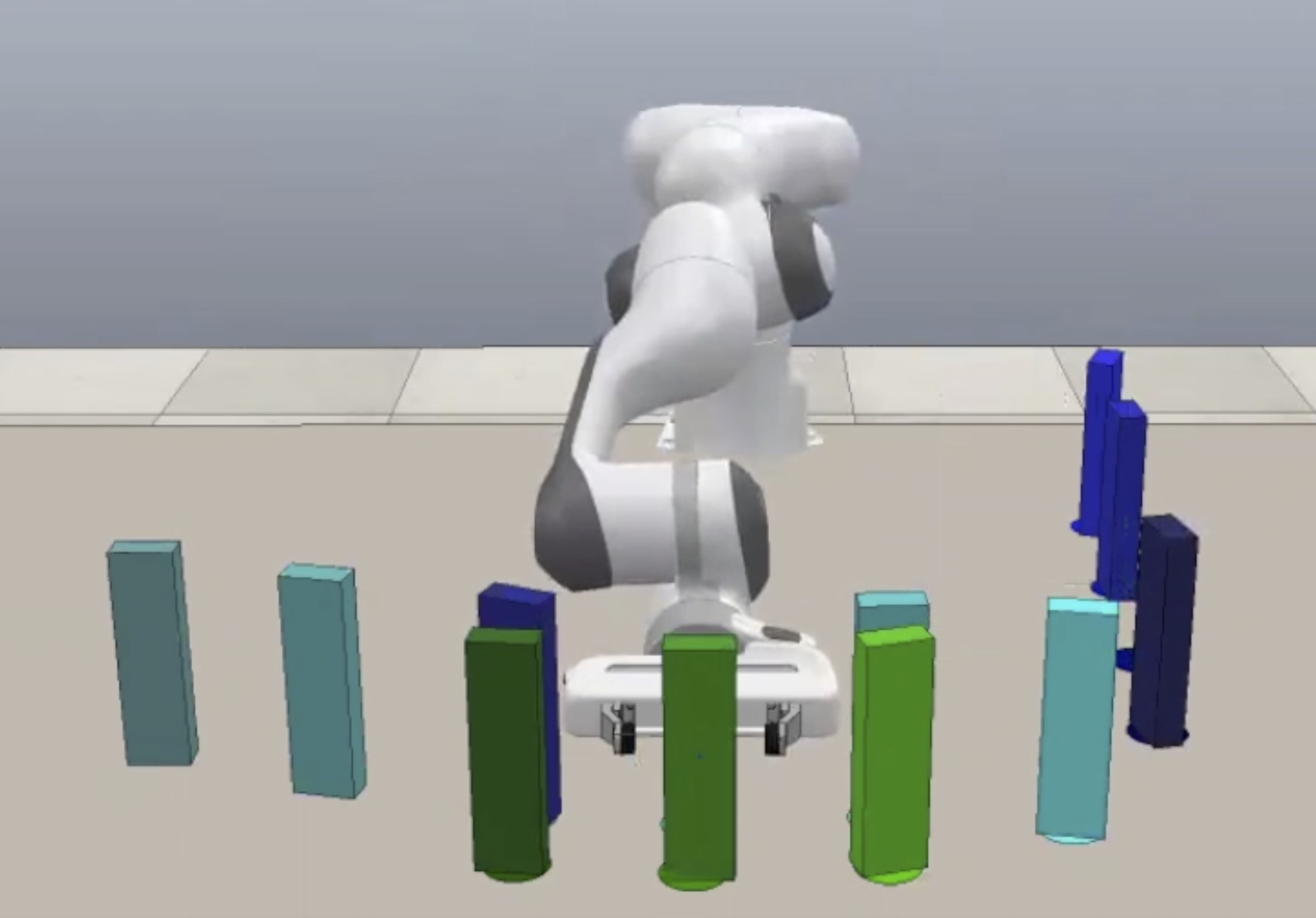}
			\label{fig:p32_6}
		}
        \subfigure[{\tt Pk-s(bcyan4,stable1)}]{
			\includegraphics[width=0.46\columnwidth]{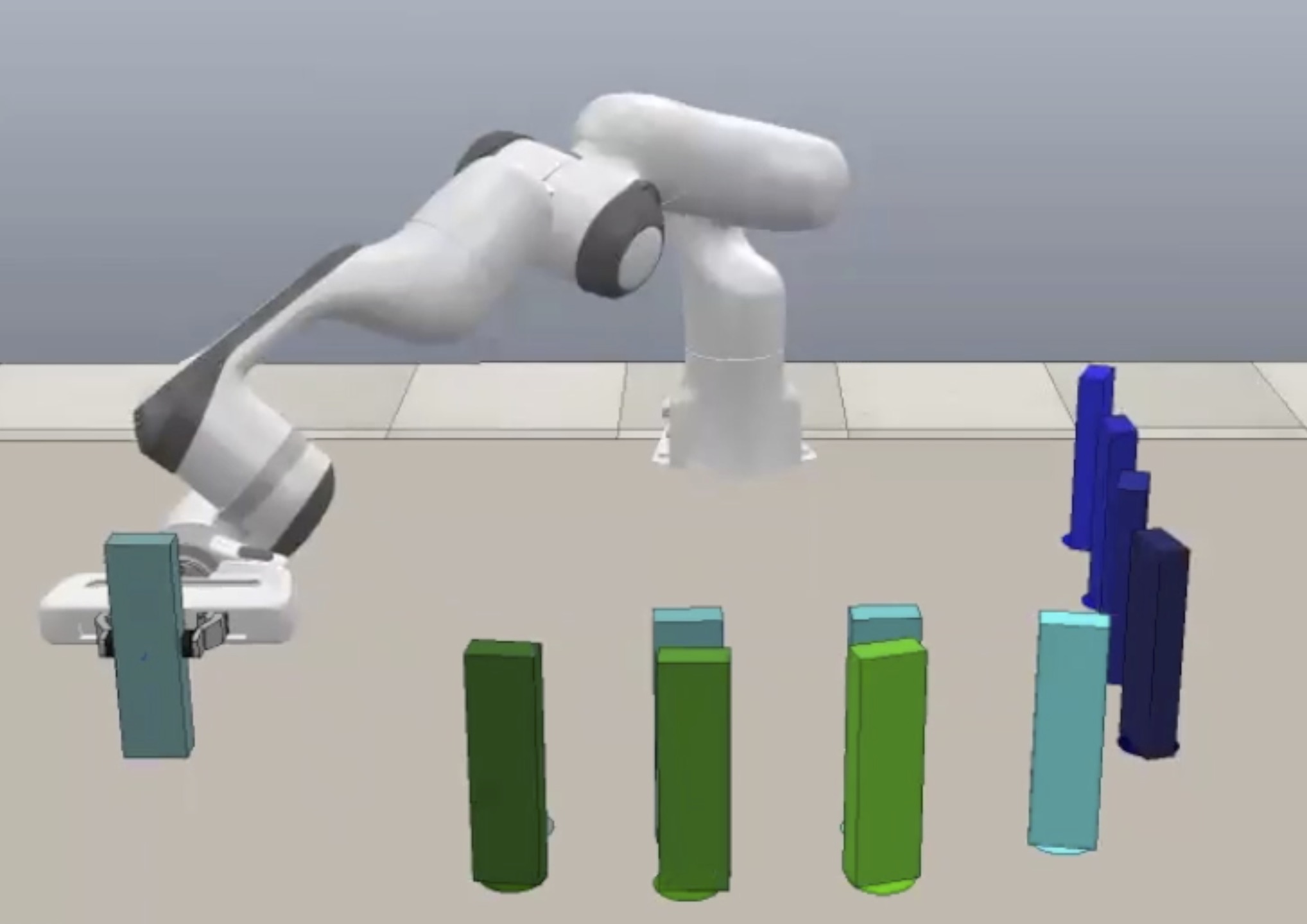}
			\label{fig:p32_7}
		}
		\subfigure[{\tt Pl-s(bcyan4,scyan4)}]{
			\includegraphics[width=0.46\columnwidth]{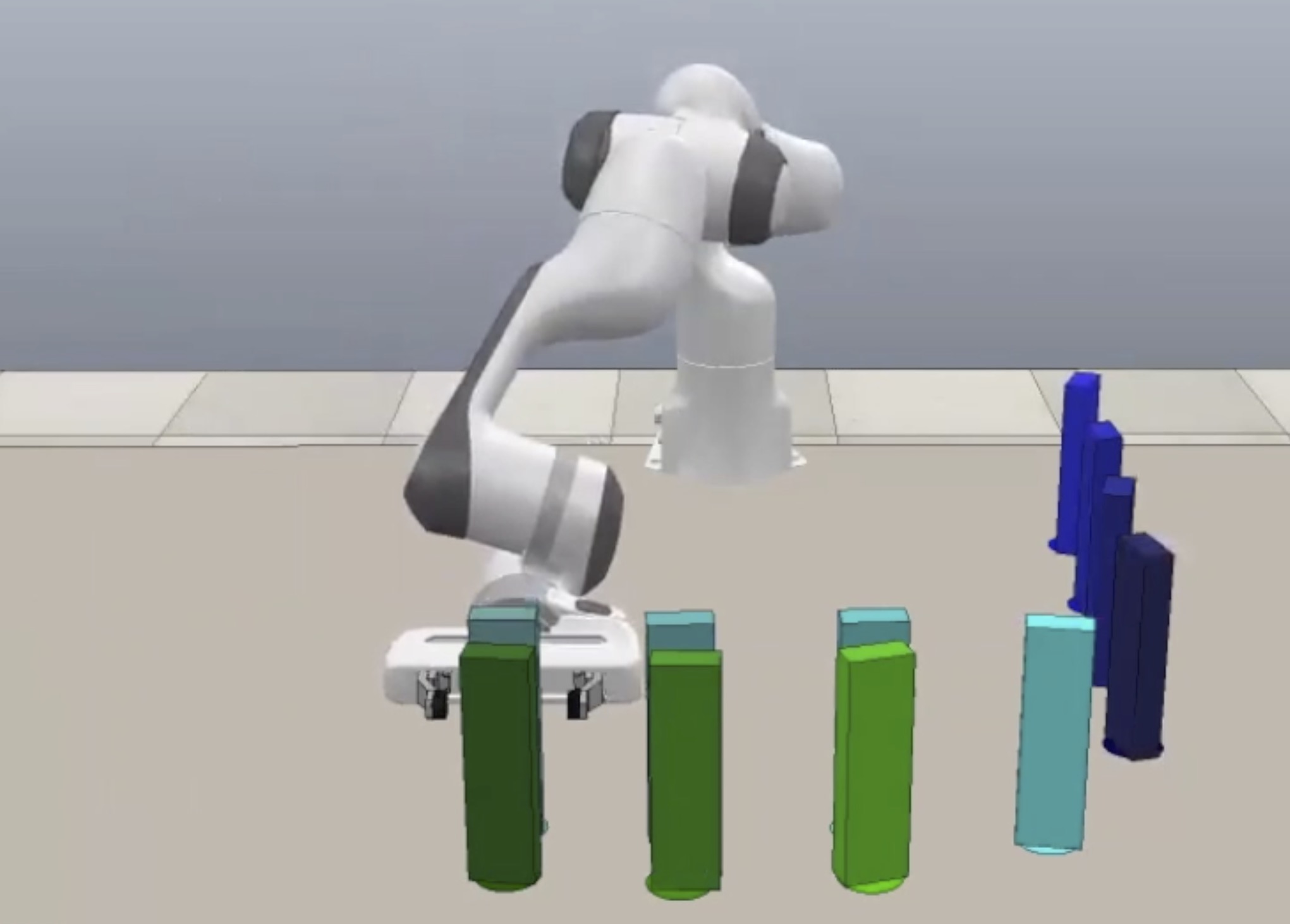}
			\label{fig:p32_8}
		}
		\caption{Snapshots of plan executions for the task of arranging green blocks. The snapshots are labelled with the corresponding grounded actions. For clarity, actions name {\tt Pick-space} and {\tt Place-space} are shorten to {\tt Pk-s} and {\tt Pl-s}. For clarity, only the grounded arguments corresponding to the manipulated blocks and the involved spaces are shown.}
		\label{fig:p32}
	\end{center}
\end{figure}

Using the U-TAMP approach, a plan comprising 46 {\tt Pick-space} and {\tt Place-space} actions was generated in 0.31 sec. The plan was executed without any failure. Fig. \ref{fig:p32} shows snapshots of the plan execution. The complete execution of the generated plan can be found at this \href{https://alejandroagostini.github.io/projects/utamp/media/demo_utamp_green_blocks_20x.mp4}{\it link}.
The performance of our U-TAMP approach is significantly better than the one reported by Garrett et al. \cite{garrett2018ffrob} for the same tasks, which require 135 sec. with an average success rate of 0.72 when applying their best performance algorithm ${\rm H_{FFRob},HA}$.   

\subsection{Task 2: Cooking Dinner for Two}
The scenario for task of cooking a dinner for two (Problem 5 in \cite{garrett2018ffrob}) is shown in Fig. \ref{fig:experiments_problem_5}. This task requires to pick the two cabbages (green blocks), clean them by placing them on the dishwasher (blue circle), cook them by placing them on the microwave (yellow circle), and finally place them on the plates (orange circles). As with the previous task, objects cannot be grasped from top or back. In order to pick the cabbages, the robot must move the blocking turnips (red blocks) and put them back in their original place (non-monotonic task). The cabbages change color when cleaned or cooked.
The task also requires to wash the cups (blue and cyan blocks) and set the table using only the blue cups. The cyan cup must be washed but it is not needed for dinner. Clean cups become transparent once they are washed.
This task requires keeping track of changes in single (e.g. cabbage-plate) and multiple (e.g. turnip-cabbage and turnip-table) sides of bounding boxes. Thus, to solve this task, we will use planning operators for object-support and object-container tasks for the planning domain definition. We also include in the domain definition the planning operators {\tt clean} and {\tt cook} defined in \cite{garrett2018ffrob}.

For the problem definition, we consider two green blocks representing cabbages, {\tt bcabbage}$i$, $i=1,2$, four red blocks representing turnips, {\tt bturnips}$i$, $i=1,..,4$, three glasses, {\tt bglass}$i$, $i=1,..,3$, where {\tt bglass3} is the cyan one, two plates {\tt plate}$i$, $i=1,2$, two places for glasses on the table {\tt table-glass}$i$, $i=1,2$, a microwave (yellow circle) {\tt microw}, and a dishwasher (light blue circle) {\tt dishw}.
In addition to the target tabletop parts predefined by the task specification to place the glasses, our algorithm generated additional tabletop parts corresponding to the places where the glasses were initially placed. These parts were automatically labelled as {\tt table}$i$, $i=1,..,3$ and used to temporally place other objects during execution.
Finally, we included in the domain definition spaces for cabbages and turnips, {\tt scabbage}$i$ and {\tt sturnips}$i$, respectively, as well as adjacent spaces to assess grasping constraints.

\begin{figure}
	\begin{center}
		\subfigure[{\tt Pk(bglass2,table1)}]{
			\includegraphics[width=0.46\columnwidth]{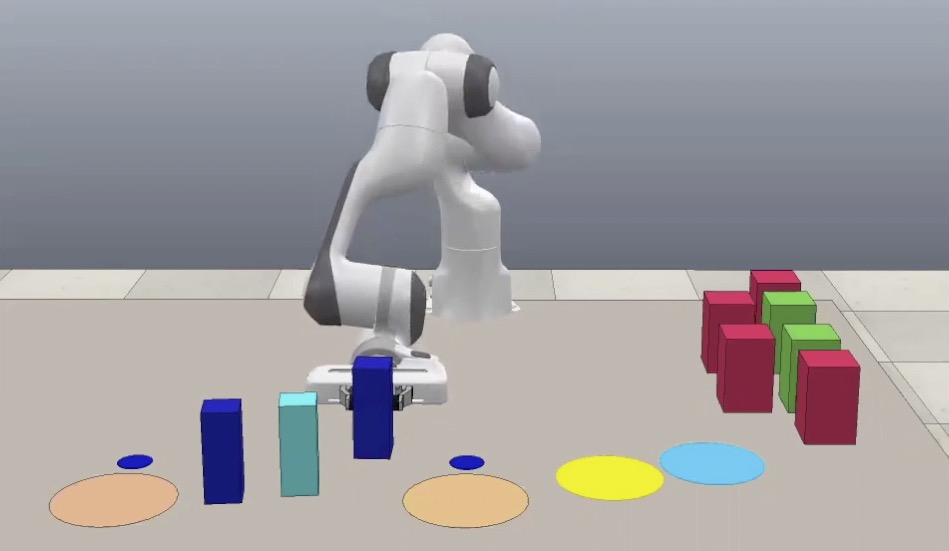}
			\label{fig:p5_1}
		}
		\subfigure[{\tt Pk(bglass1,dishw)}]{
			\includegraphics[width=0.46\columnwidth]{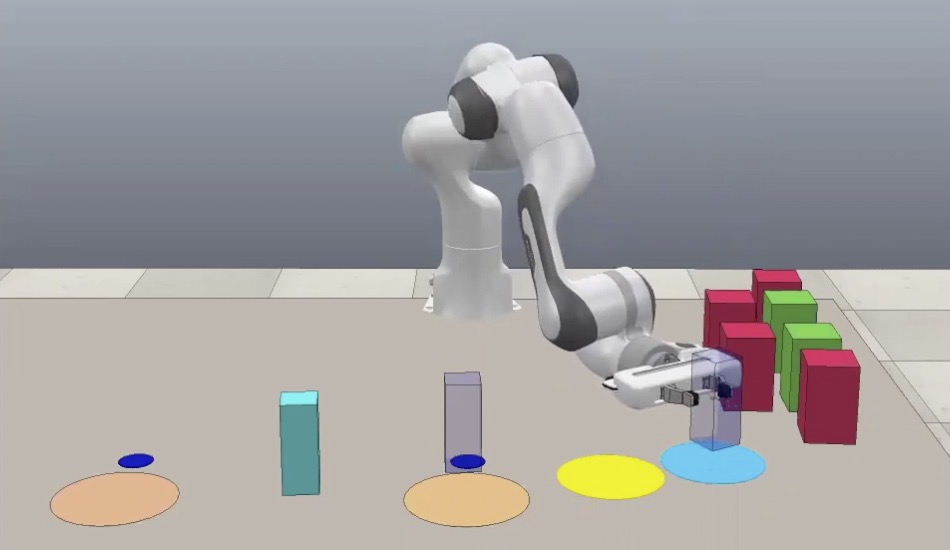}
			\label{fig:p5_2}
		}
        \subfigure[{\tt Pk-s(bcabbage2,scabbage2)}]{
			\includegraphics[width=0.46\columnwidth]{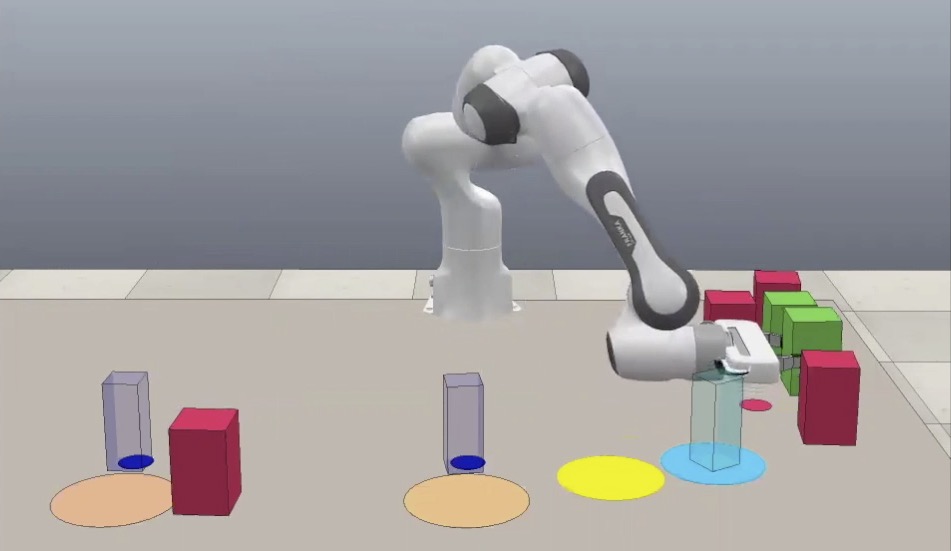}
			\label{fig:p5_3}
		}
		\subfigure[{\tt Pk(bcabbage2,dishw)}]{
			\includegraphics[width=0.46\columnwidth]{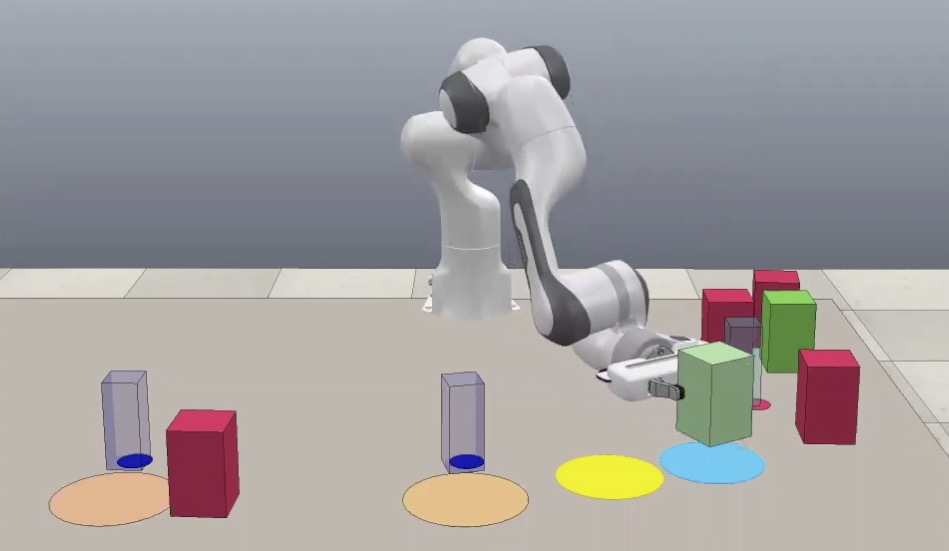}
			\label{fig:p5_4}
		}
        \subfigure[{\tt Pk(bcabbage2,microw)}]{
			\includegraphics[width=0.46\columnwidth]{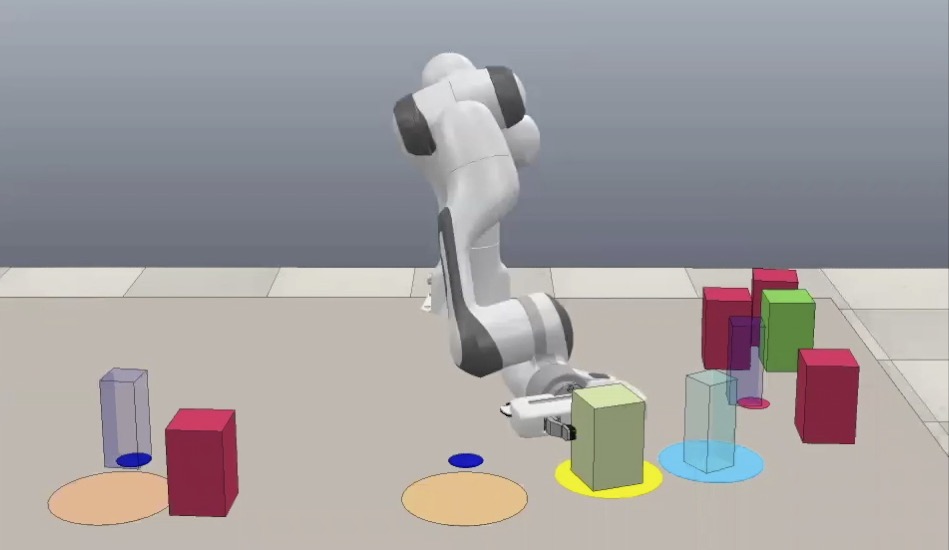}
			\label{fig:p5_5}
		}
		\subfigure[{\tt Pk-s(bcabbage1,scabbage1)}]{
			\includegraphics[width=0.46\columnwidth]{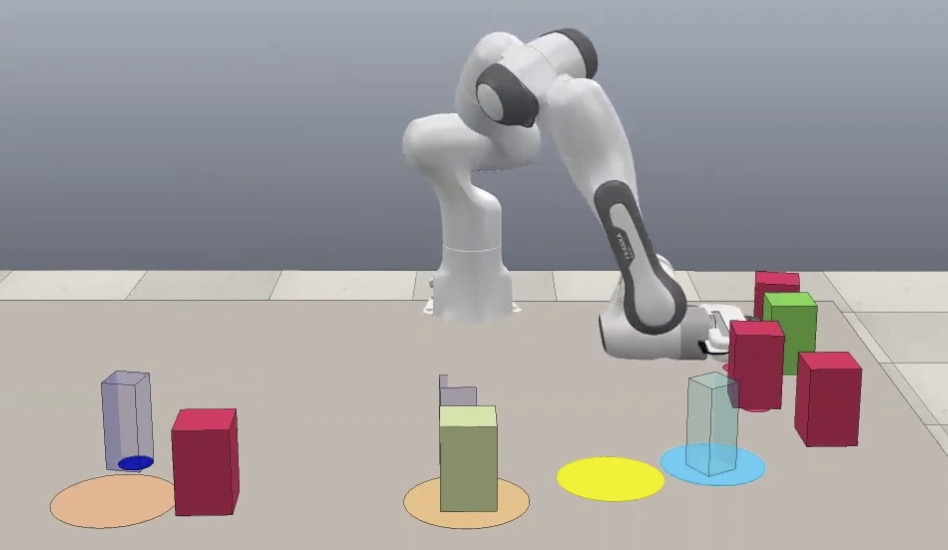}
			\label{fig:p5_6}
		}
        \subfigure[{\tt Pl(bcabbage1,plate1)}]{
			\includegraphics[width=0.46\columnwidth]{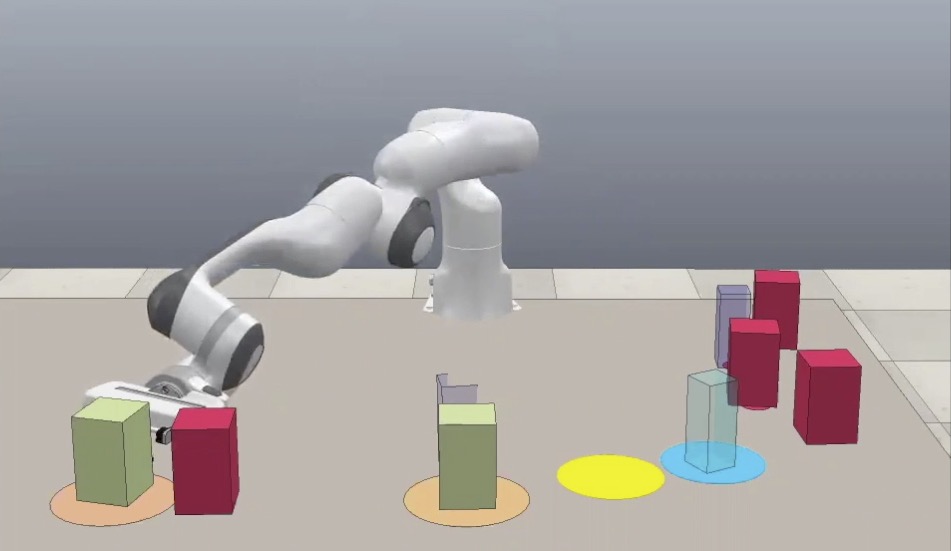}
			\label{fig:p5_7}
		}
		\subfigure[{\tt Pl-s(bturnip2,sturnip2)}]{
			\includegraphics[width=0.46\columnwidth]{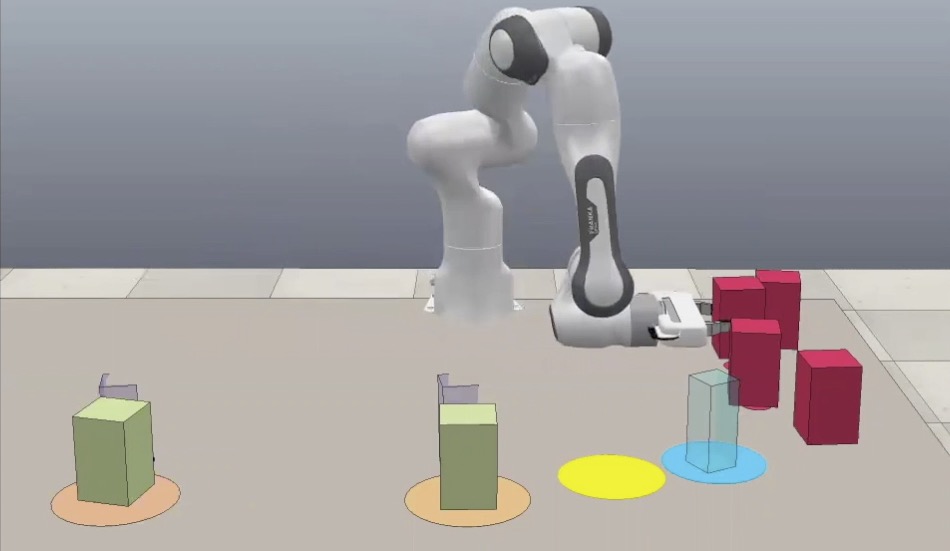}
			\label{fig:p5_8}
		}
		\caption{Snapshots of plan execution for the task of cooking dinner for two. The snapshots are labelled with the corresponding grounded actions of the task plan. For clarity, actions name {\tt Pick}, {\tt Place}, {\tt Pick-space} and {\tt Place-space} are referred as {\tt Pk-s}, {\tt Pl-s}, {\tt Pk-s} and {\tt Pl-s}, respectively. For clarity, only the manipulated objects and the support (or space) object for each action are shown.}
		\label{fig:p5}
	\end{center}
\end{figure}

The U-TAMP approach yielded a plan comprising 57 actions. Fig. \ref{fig:p5} shows snapshots of plan execution labelled with the corresponding actions in the task plan. The complete execution of the generated plan can be found at this \href{https://alejandroagostini.github.io/projects/utamp/media/demo_utamp_dinner_for_two_20x.mp4}{\it link}.
The plan was generated in 0.94 sec. and was executed without any failure. In contrast, the best result reported by Garrett et al. \cite{garrett2018ffrob} for the same tasks was a computation time for TAMP of 44 sec. with an average success rate of 0.76 using their best performance algorithm ${\rm H_{FFRob},HA}$. 

\section{CONCLUSIONS}
We proposed a TAMP approach that unifies task and motion planning into a single heuristic search. The approach is based on object-centric abstractions of TAMP constraints represented in terms of object-part interactions relevant for the task at hand. In our work, objects parts were characterized as the parts of the bounding boxes of objects, which were sufficient for a complete characterization of TAMP constraints in the block-world scenarios considered for the experimental evaluation. However, the approach can be easily extended to consider a richer set of functional object parts. 
Our approach was able to find executable solutions in challenging benchmark scenarios proposed by the state of the art requiring long-horizon executions satisfying several multi-modal constraints. 
Preliminary results suggest that the U-TAMP approach is able to significantly outperform state of the art TAMP approaches, decreasing the computation time required to generate TAMP solutions in about two order of magnitudes (Sec. \ref{sec:experiments}).

\addtolength{\textheight}{-10cm}   


%
\section*{ACKNOWLEDGMENT}
This research was funded by the Austrian Science Fund (FWF) Lise Meitner Project M2659-N38 (DOI: 10.55776/M2659) and Principal Investigator Project P36965 (DOI: 10.55776/P36965).

\bibliographystyle{IEEEtran}
\bibliography{bib.bib}

\end{document}